\documentclass[acmtog]{acmart}
\acmSubmissionID{1666}

\AtBeginDocument{%
  }

\setcopyright{acmlicensed}
\copyrightyear{2025}
\acmYear{2025}
\setcopyright{cc}
\setcctype{by}
\acmJournal{TOG}
\acmYear{2025} \acmVolume{44} \acmNumber{6} \acmArticle{226} \acmMonth{12} \acmPrice{}\acmDOI{10.1145/3763323}

\usepackage{multirow}
\usepackage{enumitem}
\usepackage{amsmath}
\usepackage{mathtools}

\usepackage{booktabs}
\usepackage{multirow}
\usepackage{makecell}

\setlist[itemize]{leftmargin=*,noitemsep}

\newcommand{\re}[1]{{\color{black}#1}}

\begin{document}

\title{BrepGPT: Autoregressive B-rep Generation with Voronoi Half-Patch}


\author{Pu Li}
\authornote{These authors contributed equally.}
\email{lipu2021@ia.ac.cn}
\affiliation{
	\institution{MAIS, Institute of Automation, Chinese Academy of Sciences, and University of Chinese Academy of Sciences}
	\country{China}	
}

\author{Wenhao Zhang}
\authornotemark[1]
\email{wenhao.zhang@ia.ac.cn}
\affiliation{
	\institution{MAIS, Institute of Automation, Chinese Academy of Sciences}
	\country{China}	
}

\author{Weize Quan}
\email{qweizework@gmail.com}
\affiliation{
	\institution{MAIS, Institute of Automation, Chinese Academy of Sciences}
	\country{China}	
}

\author{Biao Zhang}
\email{biao.zhang@kaust.edu.sa}
\affiliation{
	\institution{King Abdullah University of Science and Technology}
	\country{Saudi Arabia}	
}

\author{Peter Wonka}
\email{peter.wonka@kaust.edu.sa}
\affiliation{
	\institution{King Abdullah University of Science and Technology}
	\country{Saudi Arabia}	
}

\author{Dong-Ming Yan}
\authornote{Corresponding author: Dong-Ming Yan (yandongming@gmail.com)}
\email{yandongming@gmail.com}
\affiliation{
	\institution{MAIS, Institute of Automation, Chinese Academy of Sciences, and University of Chinese Academy of Sciences}
	\country{China}	
}

\begin{abstract}
Boundary representation (B-rep) is the de facto standard for CAD model representation in modern industrial design. The intricate coupling between geometric and topological elements in B-rep structures has forced existing generative methods to rely on cascaded multi-stage networks, resulting in error accumulation and computational inefficiency. We present BrepGPT, a single-stage autoregressive framework for B-rep generation.
Our key innovation lies in the Voronoi Half-Patch (VHP) representation, which decomposes B-reps into unified local units by assigning geometry to nearest half-edges and sampling their next pointers. Unlike hierarchical representations that require multiple distinct encodings for different structural levels, our VHP representation facilitates unifying geometric attributes and topological relations in a single, coherent format. We further leverage dual VQ-VAEs to encode both vertex topology and Voronoi Half-Patches into vertex-based tokens, achieving a more compact sequential encoding. A decoder-only Transformer is then trained to autoregressively predict these tokens, which are subsequently mapped to vertex-based features and decoded into complete B-rep models.
Experiments demonstrate that BrepGPT achieves state-of-the-art performance in unconditional B-rep generation. The framework also exhibits versatility in various applications, including conditional generation from category labels, point clouds, text descriptions, and images, as well as B-rep autocompletion and interpolation.

\end{abstract}






\begin{teaserfigure}
  \centering
  \includegraphics[width=\linewidth]{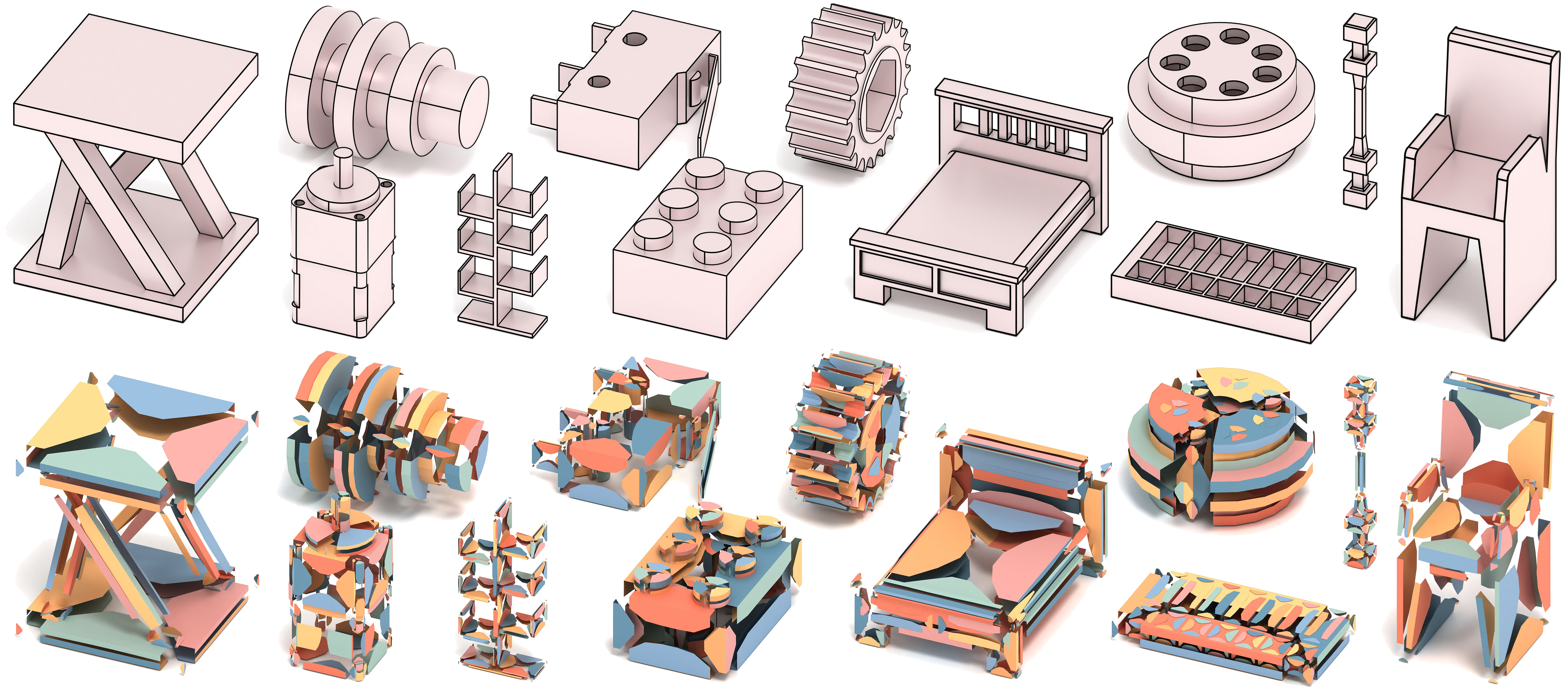}
  \caption{Top: B-rep models generated by BrepGPT. Bottom: Visualization of corresponding Voronoi Half-Patches, where distinct colors represent regions in the parametric space of each face that are geometrically closest to their respective boundary curves.} 
  \label{fig:teaser}
\end{teaserfigure}

\maketitle

\section{Introduction}

Boundary representation (B-rep) \cite{weiler1986topological} serves as the cornerstone of modern Computer-Aided Design (CAD) systems, enabling accurate modeling of intricate 3D shapes through geometrically defined faces, edges, and vertices, as well as their topological interrelations. Its hierarchical structure offers advantages compared to other representation formats, particularly in capturing smooth freeform surfaces, preserving geometric precision, and supporting robust design semantics. These features are essential in engineering workflows, where correctness, manufacturability, and structural analysis depend on exact geometry and rigid connectivity constraints. As a result, the B-rep representation facilitates a wide range of downstream tasks, including editing, analysis, simulation, etc.

Despite its importance, B-rep modeling poses significant difficulties for automatic generation. The hybrid nature of B-reps, combining continuous geometric parameters with discrete topological structures, makes the design space highly complex. Furthermore, the data structure of the B-rep representation changes for each 3D shape. B-rep models may comprise a varying number of faces, each containing boundary edges with arbitrary counts and optional inner loops. Such variability presents a challenge in directly applying deep learning techniques, which are generally tailored to inputs of fixed dimensionality. While recent generative models have demonstrated success on alternative 3D formats (\textit{e.g.}, meshes, point clouds, voxels, and implicit fields) \cite{shi2022deep,li2023generative,chen2024survey,li2024advances,sun2024recent}, they fall short of capturing the stringent accuracy and consistency requirements of CAD models. Many CAD-oriented approaches sidestep B-rep generation by instead synthesizing sketch-and-extrude construction sequences \cite{wu2021deepcad, xu2022skexgen, xu2023hierarchical, ma2024draw, ren2022extrudenet}, which fit well-established sequence generation techniques but are restricted by limited operation types and dataset availability. 

Existing B-rep generation methods are typically built on hierarchical representations, which inherently lead to multi-stage generation pipelines that suffer from stage-to-stage inconsistencies.
SolidGen \cite{jayaraman2022solidgen} emphasizes discrete data structures, sequentially defining vertices, edges, and faces via pointer networks. Its primary limitation lies in the inability to handle freeform surfaces. In contrast, BrepGen \cite{xu2024brepgen} adopts a surface-based representation similar to UV-Net \cite{jayaraman2021uv}, where face geometry is captured by uniformly sampling grid points in the parametric space. After generating boundaryless faces, the boundary edges for each face are further generated. While capable of representing freeform surfaces, its padding strategy, which ensures fixed input lengths for the diffusion module, introduces computational overhead and sensitivity to deduplication procedures. DTGBrepGen \cite{li2025dtgbrepgen} seeks to explicitly separate topology from geometry by first generating edge-face and edge-vertex adjacency relationships, followed by the detailed geometry generation using a series of diffusion models. 
\re{While this explicit separation provides interpretability, the sequential nature of multiple diffusion models increases computational complexity and inference time.}

To overcome the inherent drawbacks of hierarchical B-rep representations, we introduce a unified sequential encoding scheme that consolidates the entire B-rep structure into a single-level coherent representation. 
Concretely, we encode the continuous surface and curve geometry, as well as the discrete topology of B-reps, into vertex-based features. Each vertex feature consists of three components: 1) its 3D coordinates, 2) features capturing connectivity between vertices, and 3) information about adjacent half-edges, including their local geometric details, next-edge relationships, and inner/outer classifications within loops. 
To bridge the gap between different primitive levels, we propose the Voronoi Half-Patch (VHP) structure, which implements the third component of our vertex-based representation. The VHP is defined for each half-edge in the B-rep model, effectively decomposing the complete B-rep information into local half-edge-centric units. This decomposition occurs in two ways: geometrically, we partition each surface in its parametric domain using Voronoi diagrams, assigning the nearest surface region to its corresponding half-edge; topologically, we encode the next-edge relationships through sampling along adjacent half-edges, enabling the reconstruction of face boundary loops through sequential half-edge traversal.
The name Voronoi Half-Patch reflects its key characteristics: \textit{Voronoi} refers to the parametric space partitioning mechanism, \textit{Half} denotes its basis in the half-edge data structure, and \textit{Patch} signifies its representation of both curve geometry and associated surface regions.

This vertex-based representation offers several key advantages: 1) eliminating the need to repeatedly generate shared edges, 2) supporting freeform surfaces, 3) enabling single-stage training for \re{deep generative networks}, in contrast to the multi-stage pipelines of prior approaches, and 4) naturally providing a vertex-based tokenization mechanism for B-reps, facilitating autoregressive generation without redundant padding and deduplication. Technically, we utilize two separate VQ-VAEs, each with a graph convolutional encoder that consolidates vertex connectivity or VHP inputs into vertex embeddings, which are later quantized into discrete codes. This results in a compact 11-dimensional token sequence for each vertex feature. We train a GPT-style Transformer to autoregressively generate these token sequences. 
During the decoding phase, we extract two types of features from the sequence: connectivity features and VHP features. The former \re{defines edge-vertex topology}, while the latter determines \re{face-edge topology} and specifies geometric definitions of curves and surfaces. The integration of these components yields complete, high-fidelity B-rep models.
To summarize, the core contributions of this paper include the following:
\begin{itemize}
    \item A novel Voronoi Half-Patch (VHP) representation that enables uniform local encoding of B-rep models, in contrast to previous hierarchical approaches.
    \item A vertex-based tokenization scheme built upon VHP to achieve a concise sequential representation of B-rep models.
    \item Formulating B-rep generation as a sequence generation task, demonstrating its effectiveness in unconditional, multi-modal conditional, autocompletion, and interpolation applications.
\end{itemize}
\section{Related Work}


\paragraph{Constructive solid geometry}
Constructive Solid Geometry (CSG) constructs sophisticated CAD models by applying Boolean operations to elementary primitives such as cuboids and spheres, yielding a compact and interpretable tree-based representation. Historically, CSG reconstruction approaches were paired with program synthesis \cite{du2018inversecsg, nandi2017programming, nandi2018functional}, later significantly benefiting from learning-based techniques to increase precision \cite{sharma2018csgnet, ellis2019write, tian2019learning}. Recent literature further explores unsupervised methodologies and specialized parametric primitives, alleviating the dependence on explicit supervision from ground-truth CSG trees \cite{kania2020ucsg, chen2020bsp, ren2021csg, yu2022capri, yu2023d, jones2022plad, ganeshan2023improving}. Nevertheless, neural CSG reconstructions often contain redundant primitives, reflecting solution ambiguity, and exhibit limited expressiveness compared to the feature-based modeling paradigm. While converting CSG models into B-reps is feasible through a geometry kernel, such conversions can introduce undesirable complexities and geometric artifacts. Instead of relying on intermediate CSG representations, our approach learns to directly generate B-reps seamlessly compatible with CAD editing workflows.

\paragraph{Sketch-and-extrude modeling sequence}
Framing CAD generation as sequences of sketch-and-extrude operations has become a popular choice, which leverages construction histories from parametric CAD files. Early works demonstrated the feasibility of generating 2D engineering sketches \cite{willis2021engineering, para2021sketchgen, ganin2021computer, seff2022vitruvion}. Building on DeepCAD’s introduction of a large-scale 3D CAD modeling sequence dataset \cite{wu2021deepcad}, subsequent studies have progressively incorporated advanced techniques (\textit{e.g.}, autoregressive frameworks and diffusion models) to enhance the quality and diversity of generation results \cite{xu2022skexgen, xu2023hierarchical, wang2024vq, li2025revisiting}. CAD model reconstruction from inputs of various forms has also been explored through both supervised and unsupervised strategies \cite{xu2021inferring, lambourne2022reconstructing, uy2022point2cyl, ren2022extrudenet, li2023secad,li2024sfmcad,uwimana2025segmentation}. Recently, there has been growing interest in generating full 3D CAD models from ambiguous and underspecified modalities such as images and natural language descriptions \cite{you2024img2cad, chen2024img2cad, sadil2024text2cad, xu2024cad}. However, current sequence-based methods predominantly focus on sketch-and-extrude operations, lacking support for advanced CAD commands such as loft, sweep, fillet, and chamfer. The absence of explicit topological structures
further constrains their capability to produce intricate shapes. Moreover, the limited size of the dataset with operation sequences, compared to the existing B-rep dataset (5 times larger) \cite{koch2019abc}, hinders the scalability of training for sequence-based methods.

\paragraph{Boundary representation}

B-rep is widely recognized as a fundamental format in modern CAD applications. Structurally, a B-rep model comprises geometric entities
as well as topological elements,
which collectively describe complex solid structures through adjacency and connectivity. Representing B-reps as graphs, initially introduced by \cite{ansaldi1985geometric}, has significantly influenced recent advances in applying graph-based neural architectures to diverse shape analysis tasks, notably classification and segmentation \cite{cao2020graph, jayaraman2021uv, willis2022joinable, lambourne2021brepnet, bian2024hg, jones2021automate, jones2023self}. For instance, UV-Net \cite{jayaraman2021uv} employs a face adjacency graph and conducts CNN-based feature extraction on parametric surfaces and curves, combined through graph convolution. Similarly, BRepNet \cite{lambourne2021brepnet} integrates custom convolutional operations directly onto B-rep primitives.
Other graph-based frameworks have emerged to comprehensively model intricate interactions among varying primitive types, accommodating finer details \cite{bian2024hg, jones2021automate, willis2022joinable}. 

B-rep generation task seeks to simultaneously construct both geometric primitives and topological relationships of CAD solids, which is proven challenging due to their intertwined nature. Earlier research primarily focused on geometric construction of parametric primitives \cite{wang2020pie, sharma2020parsenet, li2023surface}, often assuming rigid topological templates \cite{smirnov2021patches}. Later work \cite{willis2021engineering} expanded the modeling capacity to support flexible topological configurations for sketches. The wireframe, as a subset of the B-rep that omits surface geometry, has been investigated as a generative target as well \cite{ma2024generating, ma2025clr}. Recent methods have also made advances in B-rep model reconstruction from varied inputs such as 2D wireframe drawings \cite{wang2022neural} and point clouds \cite{guo2022complexgen, liu2024split, liu2024point2cad}. 

SolidGen \cite{jayaraman2022solidgen} achieved direct B-rep synthesis through a sequential pipeline that autoregressively models vertices, edges, and faces using pointer networks, capturing topological dependencies explicitly. However, its reliance on prismatic constraints limits generalization. BrepGen \cite{xu2024brepgen} leverages multi-stage diffusion models to generate geometric elements hierarchically, from faces to edges to vertices, recovering topology by merging duplicated elements. DTGBrepGen \cite{li2025dtgbrepgen} proposes a framework that decouples topology and geometry generation, first predicting adjacency relationships then synthesizing coordinates and B-spline control points, though this multi-phase approach remains prone to error propagation. \re{Most recently, HoLa \cite{liu2025hola} introduces a holistic latent representation unifying geometric and topological information through a neural intersection network.} However, all diffusion-based approaches \cite{xu2024brepgen,li2025dtgbrepgen,liu2025hola} suffer from a fundamental limitation: they require input padding to match fixed-length requirements, conflicting with B-rep's flexible nature and introducing training overhead and inference fragility due to required deduplication. \re{In contrast, our approach employs a half-edge-centric VHP representation with autoregressive synthesis.}

\paragraph{GPT-style 3D shape generation}

The autoregressive framework has demonstrated its potential in the context of mesh modeling, where meshes can be viewed as a simplified variant of B-rep models composed of planar faces and linear edges. 
Drawing inspiration from the success of generative language models \cite{radford2018improving, radford2019language, brown2020language}, MeshGPT \cite{siddiqui2024meshgpt} reformulates the problem by first encoding mesh triangles into discrete tokens via VQ-VAE \cite{van2017neural}, and then generating them sequentially using a decoder-only Transformer \cite{vaswani2017attention}. This paradigm has sparked follow-up research on more compact mesh tokenizations and scalable architectures \cite{chen2024meshanything, chen2024meshanythingv2, hao2024meshtron, weng2024scaling, lionar2025treemeshgpt}, enabling the generation of complicated meshes with higher face counts.

In the domain of CAD model generation, existing GPT-style methods \cite{xu2022skexgen, xu2023hierarchical,wang2025cad,xu2024cad} are primarily based on command sequences, yet they are constrained by non-trivial challenges such as limited operation types and small-scale datasets. 
SolidGen \cite{jayaraman2022solidgen,10.1145/3721238.3730661} adopts an autoregressive framework for direct B-rep generation, but relies on pointer networks rather than a GPT-based decoder. Stitch-A-Shape \cite{10.1145/3721238.3730661} employs an autoregressive approach solely during vertex generation, while lacking comprehensive face geometry encoding. To the best of our knowledge, this paper is the first work to introduce a tokenized representation of full B-rep models and employ a GPT-style generative model that performs direct B-rep generation. 

\section{Voronoi Half-Patch Representation}

\begin{figure}[!t]
\centering
  \includegraphics[width=1\linewidth]{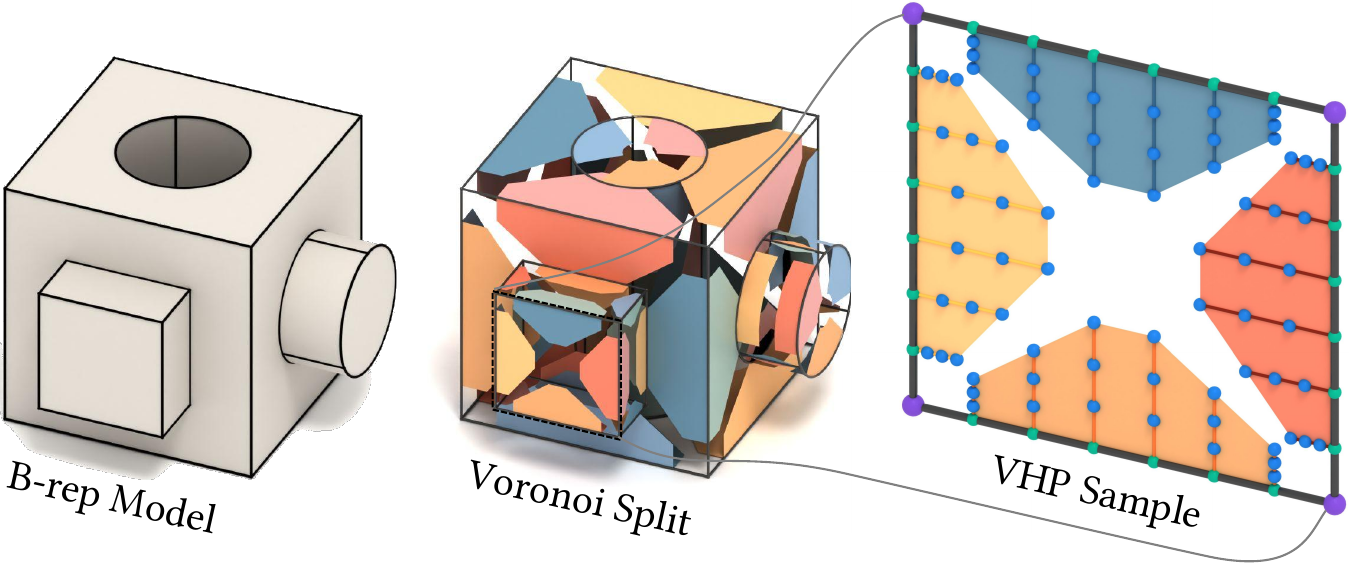}   
  \caption{Geometric sampling strategy for Voronoi Half-Patches (VHP). Left: Input B-rep model. Middle: Voronoi partitioning in parametric space. Right: \re{Geometric sampling} within Voronoi regions, showing vertices (purple), curve samples (green), and surface samples (blue) with sampling paths.}
  \label{fig:vhp_geometry}
\end{figure}

In this section, we introduce the Voronoi Half-Patch (VHP) representation, which decomposes B-rep models into local half-edge-centric units that encapsulate both geometric attributes and topological relations. This representation leverages Voronoi diagrams to subdivide surface geometry into multiple regions, each uniquely associated with an edge based on spatial proximity. Each edge independently encodes its adjacent surface geometry alongside its intrinsic curve geometry. 
Topologically, our representation employs the half-edge structure, encoding the connectivity through directed half-edge cycles and distinguishing between inner and outer loops. 
In the following subsections, we detail the geometric and topological aspects of the VHP representation.

\subsection{Half-Patch Geometry}

\paragraph{Curve geometry}

Similar to UV-Net \cite{jayaraman2021uv} and BrepGen \cite{xu2024brepgen}, our method represents the geometry of an edge by uniformly sampling 3D points along its parametric curve. Each edge is defined by a parametric function that maps a 1D parameter $u$ to a 3D point in space.
Specifically, we sample a series of 3D points along the parameter domain of each edge, excluding the start and end points, which results in a 1D array of 3D coordinates, denoted as $C \in \mathbb{R}^{N_c \times 3}$, where $N_c$ is the number of sampled points. 
This array serves as the geometric curve feature, encoding the spatial structure of an edge.

\paragraph{Surface geometry}

In recent B-rep generation methods \cite{xu2024brepgen, li2025dtgbrepgen}, a bounded face is represented by trimming the base parametric surface with its boundary curves.
However, since faces can be bounded by varying numbers of edges, previous methods typically adopt repeat padding to achieve fixed-length representations, which introduces unnecessary computational overhead during training. 
To overcome this problem, we introduce a novel perspective: instead of treating a face as a single unit, we divide it into regions naturally associated with each boundary edge — like splitting a pie into slices. Under the manifold assumption, each edge is shared by exactly two faces, with each half-edge responsible for encoding one portion of its adjacent face. This automatically transforms the variable-length face representation into fixed-length half-edge-centric descriptions, while preserving the complete surface information. We term these edge-associated surface regions \textit{half-patches}.

Specifically, to define these half-patches, we leverage the idea of Voronoi diagrams, which partition space into regions based on the closest distance to a set of predefined sites — in this case, the edges, where distances are computed in the parametric domain of each face. As shown in Figure \ref{fig:vhp_geometry}, each face is thus subdivided into multiple half-patches, where every portion of the face geometry is assigned to the closest half-edge. This partitioning ensures that every half-edge encodes a meaningful fraction of its associated surface, capturing the local geometric details without unnecessary overlap.

For each of the $N_c$ sampled points along the curve, we further sample $N_s$ geometric points evenly along its corresponding normal direction. By combining the curve geometry with its associated surface geometry, a half-patch can be represented as $P \in \mathbb{R}^{N_c \times N_s \times 3}$, providing a compact yet comprehensive spatial representation of the local surface geometry and curve geometry.

\subsection{Half-Edge Topology}
\begin{figure}[!t]
\centering
  \includegraphics[width=1\linewidth]{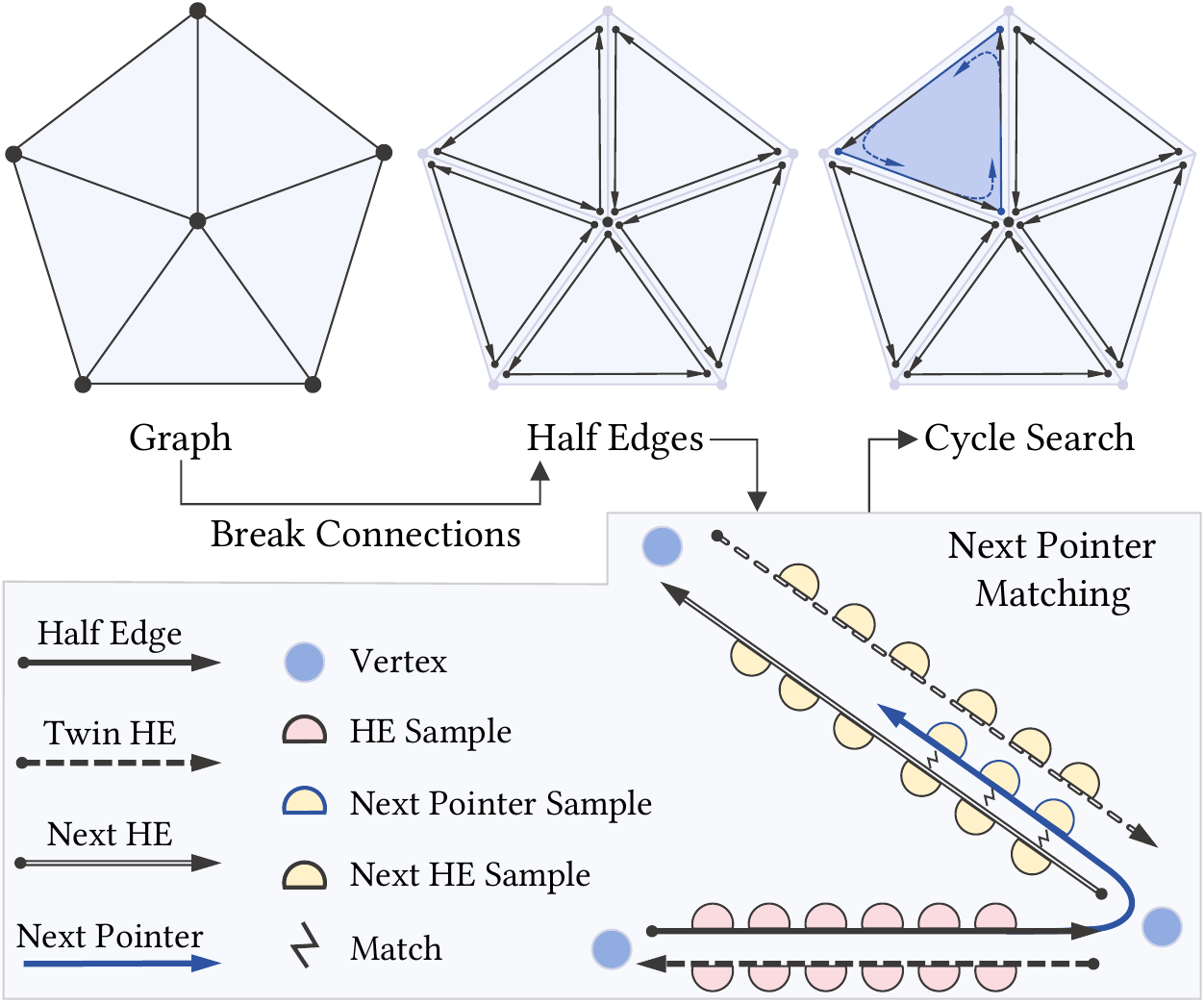}   
  \caption{Half-edge next pointer sampling illustration. Left: Vertex-edge graph structure. Middle: Decomposition into directed half-edges. Right: Next pointer establishment through strategic sampling, where terminal samples from adjacent half-edges are stored in the current VHP.}
  \label{fig:next_relationship}
\end{figure}
\paragraph{Next-edge relationship}

The VHP representation decomposes B-rep faces into multiple VHPs, requiring a reconstruction process to identify which VHPs belong to the same face. Since VHPs are defined on half-edges, they can be connected in sequence to form cycles, with each cycle representing a complete face boundary loop. The key to this reconstruction lies in determining the \textit{next} pointers for each VHP, as these pointers enable traversal through successive VHPs until returning to the starting point, thereby identifying complete faces.
SpaceMesh \cite{shen2024spacemesh} tackles a similar task via local neighborhood ordering, a method suited to manifold meshes with simple graph connectivity. However, B-reps often include self-loops and parallel edges, where such assumptions fail. To address this, we introduce a sampling-based approach that leverages continuous geometric signals to infer discrete topological adjacency, enabling robust pointer assignment in general B-rep structures.

As illustrated in Figure \ref{fig:next_relationship}, beyond sampling the current half-edge's VHP curve and surface geometry, we sample points along the initial segment of the next half-edge. \re{While we use the term "next pointers" for simplicity, these are geometric samples rather than direct topological links.} Specifically, we select the nearest $N_n$ curve sample points from the $N_c$ samples on the next half-edge, typically setting $N_n=N_s$ in practice. These next pointer samples are stored within the current VHP, resulting in a data representation in $\mathbb{R}^{(N_c+1) \times N_s \times 3}$ for each half-edge that incorporates both half-patch and next-edge samples. During reconstruction, \re{instead of direct pointer following,} we formulate the next pointer assignment as a linear assignment problem at each vertex, matching incoming and outgoing half-edges by minimizing the total Euclidean distance between next pointer samples and their corresponding candidate half-edge samples.

\paragraph{Inner/outer loop label}

Each half-edge necessarily belongs to a wire that, when closed, constitutes either an inner loop or an outer loop. Therefore, each half-edge can also be classified as either inner or outer, which can be conveniently encoded using a binary label.

By integrating the geometric details of curves and surfaces, along with topological information, including the next-edge relationship and inner/outer loop classification, we obtain a complete VHP, expressed as $\mathcal{V} \in \mathbb{R}^{(N_c+1) \times N_s \times 3 + 1}$. Together with coordinates and connectivity relationships of vertices, this structure enables the complete reconstruction of a 3D B-rep model.

\begin{figure}[!t]
\centering
  \includegraphics[width=1\linewidth]{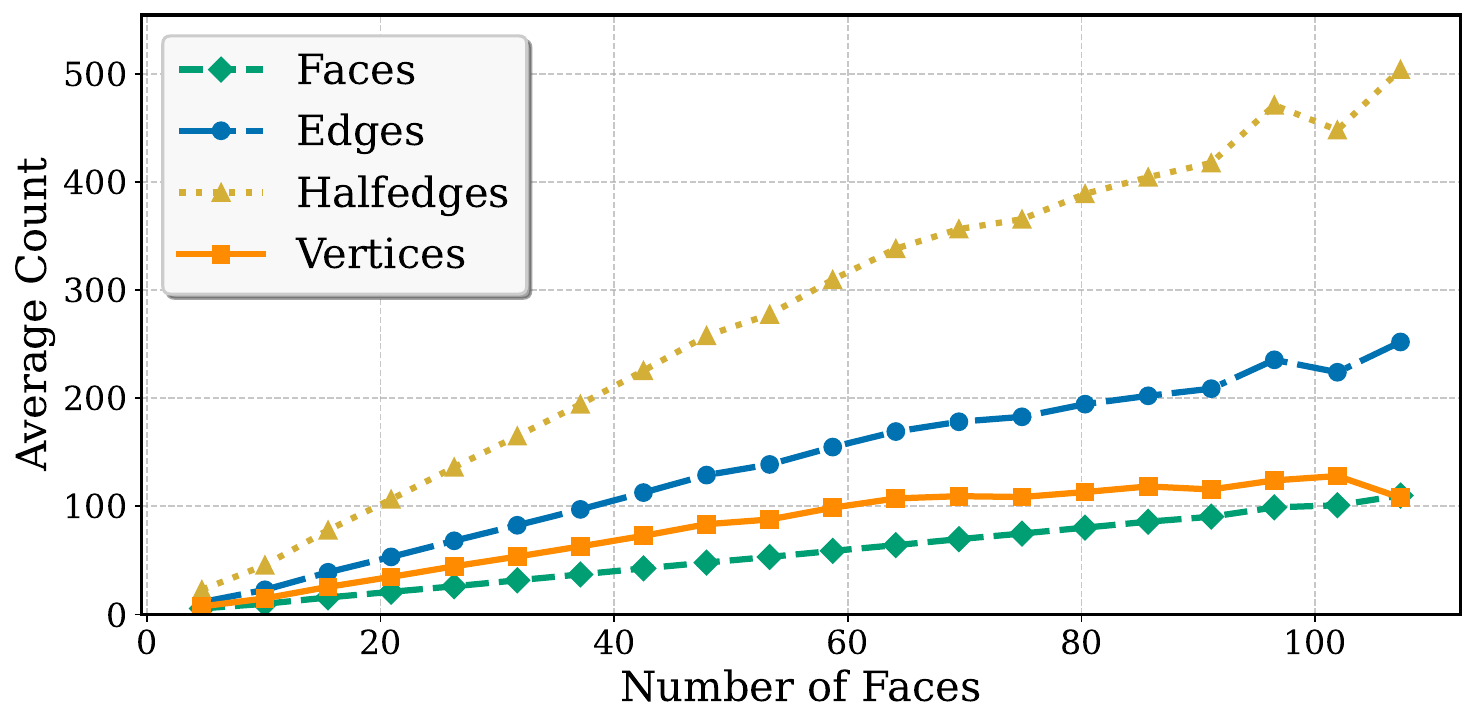}   
  \caption{Distribution of topological elements in 2,000 B-rep models from the DeepCAD dataset. For models with equivalent face counts, the number of edges consistently exceeds the vertex count, while half-edges show the highest frequency. As the face count surpasses 100, the half-edge count approaches approximately five times the vertex count.}
  \label{fig:topology_analysis}
\end{figure}
\section{Method}

\begin{figure*}[!t]
\centering
  \includegraphics[width=1\linewidth]{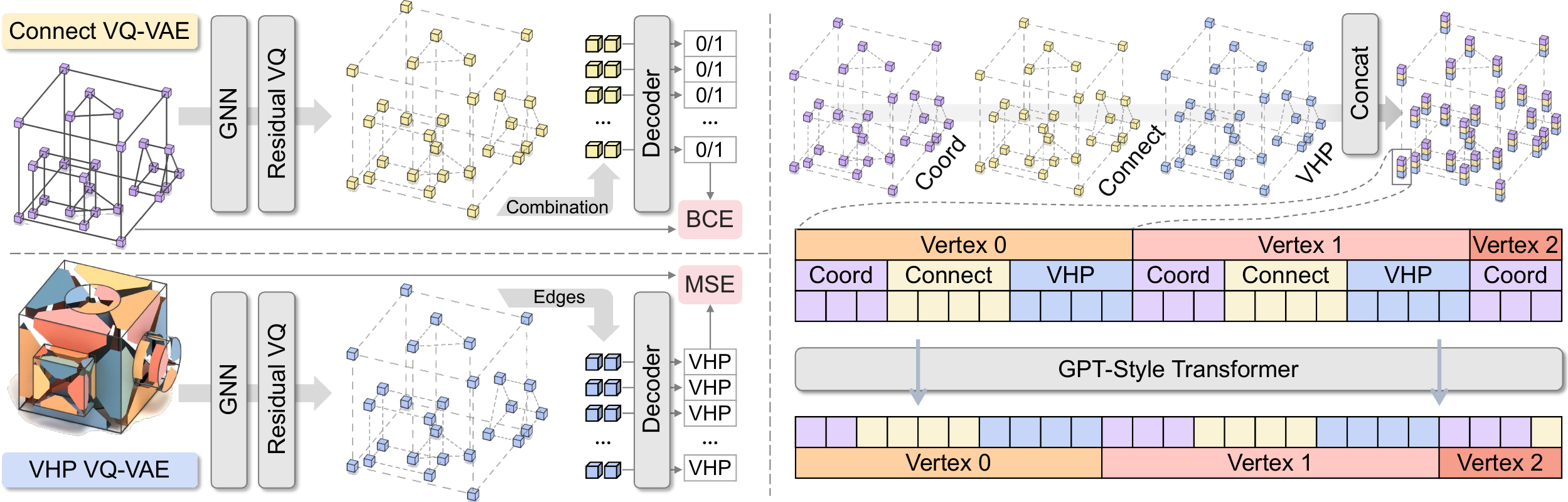}   
  \caption{Pipeline overview of BrepGPT. The Connect VQ-VAE encodes topological relationships between B-rep vertices through pairwise connectivity classification. The VHP VQ-VAE encodes geometric information within Voronoi Half-Patches through MSE regression, taking feature vectors from the start and end vertices of each directed half-edge. The GPT-style Transformer performs autoregressive generation by concatenating vertex coordinates, connectivity, and VHP tokens in sequence.}
  \label{fig:pipeline}
\end{figure*}

In this section, we first introduce our vertex-based tokenization strategy that enables a compact sequence representation while capturing full geometric and topological attributes of B-reps. We then present the dual VQ-VAEs that facilitate this compact B-rep sequence representation, followed by our GPT-style autoregressive network for sequence generation.

\subsection{Vertex-based B-rep Tokenization}

While directly applying VHP sequences for B-rep generation is technically feasible, the half-edge-based representation results in unnecessarily long sequences. For autoregressive generation tasks, prior works \cite{chen2024meshanythingv2,weng2024pivotmesh,hao2024meshtron} have established that compact and well-structured token sequences are critical for efficient model training and scaling.
In seeking a more compact sequence representation for B-rep models, we draw inspiration from the fundamental Euler-Poincaré formula for closed orientable surfaces:
\begin{equation}
V - E + F = 2(1-G),
\end{equation}
where $V$, $E$, and $F$ denote the number of vertices, edges, and faces, respectively, and $G$ represents the genus. By rearranging the equation to $V = E - F + 2(1-G)$, we can observe that for typical CAD models where $G \geq 0$ and $F \geq 2$, the term $2(1-G)$ is at most 2, and $-F$ contributes a value of at least -2. Consequently, $V$ is less than or equal to $E$, establishing that the number of vertices never exceeds the number of edges in typical B-rep models.
To validate this relationship empirically, we analyzed 2,000 randomly sampled B-reps from the DeepCAD dataset \cite{wu2021deepcad}. Figure \ref{fig:topology_analysis} demonstrates that for B-reps with equivalent face counts, the vertex count consistently remains lower than both edge count and half-edge count. \re{Our empirical validation confirms that the $V \leq E$ relationship holds universally across all examined models in our dataset.} This observation motivates our design choice of vertex-based feature sequences for our B-rep representation, offering a more compact tokenization scheme \re{over the more straightforward half-edge-based approach, thereby reducing generation errors in autoregressive modeling}.

Specifically, a B-rep model can be represented as a sequence of vertex features:
\begin{equation}
B \coloneqq (\mathbf{v}_1, \mathbf{v}_2, \ldots, \mathbf{v}_N),
\label{eq:sequence}
\end{equation}

where \(N\) denotes the number of vertices. Each vertex feature comprises three components:
\begin{enumerate}
    \item \textit{Vertex coordinates}: The spatial position $(x, y, z)$ of the vertex.
    \item \textit{Connectivity features}: A feature vector capturing the topological relationships between vertices.
    \item \textit{VHP features}: A feature vector derived from the Voronoi Half-Patch (VHP) representation, aggregating geometric and topological information from all connected half-edges where the vertex serves as either a start or end point.
\end{enumerate}

Therefore, each vertex feature can be expressed as
\begin{equation}
  \mathbf{v}_i = (x_i, y_i, z_i, \mathbf{f}^c_{i}, \mathbf{f}^v_{i}),  
\end{equation}
where \(\mathbf{f}^c_{i}\) and \(\mathbf{f}^v_{i}\) represent the connectivity and VHP features, respectively. These features are processed into discrete tokens through two different methods: the coordinates are directly quantized to 7-bit integers within $[0, 128)$, while connectivity and VHP features are encoded through VQ-VAE into 4 tokens each, resulting in an 11-dimensional token sequence per vertex (3 coordinate tokens + 4 connectivity tokens + 4 VHP tokens). To provide an intuitive understanding of token composition: coordinate tokens alone represent spatial geometry similar to point clouds; coordinate tokens combined with connectivity tokens enable graph-like representations; finally, the complete set of coordinate, connectivity, and VHP tokens enables full B-rep representations. Figure~\ref{fig:pipeline} demonstrates the tokenization and generation process of B-rep sequences in our framework. The detailed design and implementation of the dual VQ-VAEs are presented in Section \ref{subsec:quantized_embeddings}.

\subsection{Learning Quantized Vertex Embeddings} \label{subsec:quantized_embeddings}

Unlike existing autoregressive mesh generation methods \cite{siddiqui2024meshgpt, chen2024meshanything} that use a single VQ-VAE to encode triangular mesh topology and vertex coordinates, our approach employs dual VQ-VAEs to separately encode vertex connectivity and VHP features. This separation enables accurate reconstruction of both topological relationships between vertices and the non-linear, non-planar characteristics of B-rep surfaces and curves. Each VQ-VAE follows an encoder-decoder structure and employs residual vector quantization (RQ) \cite{martinez2014stacked} to quantize the vertex embeddings into discrete codebooks.

\paragraph{VQ-VAE for connectivity} \label{sec:vq_vae_connect}

To apply graph convolutions for connectivity learning, we represent vertices as nodes within a graph, where each node is described by its spatial coordinates, and the connectivity between vertices is defined with undirected edges. The encoder $E_{\text{conn}}$ utilizes multiple SAGEConv \cite{hamilton2017inductive} layers to extract a feature vector for each vertex, fusing the underlying features among adjacent vertices into learned embeddings: 
\begin{equation}
    \text{E}_{\text{conn}}(\mathcal{G}^c) = \mathbf{Z}^{c} = (z^{c}_1,z^{c}_2,\ldots,z^{c}_N),
\end{equation}
where $\mathcal{G}^c$ denotes the input connectivity graph, and $\mathbf{Z}^{c}$ represents the set of learned embeddings, each  defined as $z^{c}_i \in \mathbb{R}^{N_z}$ for the $i$-th vertex.

Based on the obtained embeddings, we employ RQ to learn quantized features. Instead of assigning each embedding a single quantized code, we enhance expressiveness by using a stack of $D_c$ codes from a set of distinct codebooks $\mathcal{B}^{c}$. 
Formally, a latent embedding $z^{c}_i$ can be represented as follows,
\begin{equation}
    \mathbf{t}^c_{i} = (t^c_{i,1}, t^c_{i,2}, \ldots, t^c_{i,D_c}) = \text{RQ}(z^c_i; \mathcal{B}^{c}, D_c),
\end{equation}
where $\mathbf{t}^c_{i}$ is a stack of tokens, with $t^c_{i,d}$ denoting the quantized code at depth $d$. 

In the decoding stage, we adopt a ResNet-based \cite{he2016deep} decoder, which is composed of a linear projection layer and a series of residual blocks. It takes concatenated embeddings of two vertices as input and outputs the predicted connectivity between them:
\begin{equation}
\hat{y}_{ij}^{c} = \text{D}_{\text{conn}} ( z_i^{c} \oplus z_j^{c} ),
\end{equation}
where $\hat{y}_{ij}^{c}$ presents the predicted result and $\oplus$ denotes the concatenation operation.
The loss function consists of two main components: the binary cross-entropy loss guiding accurate connectivity classification, and the vector quantization loss that includes a codebook loss and a commitment loss, together promoting efficient codebook utilization and representation stability.

\begin{figure}[!t]
\centering
  \includegraphics[width=1\linewidth]{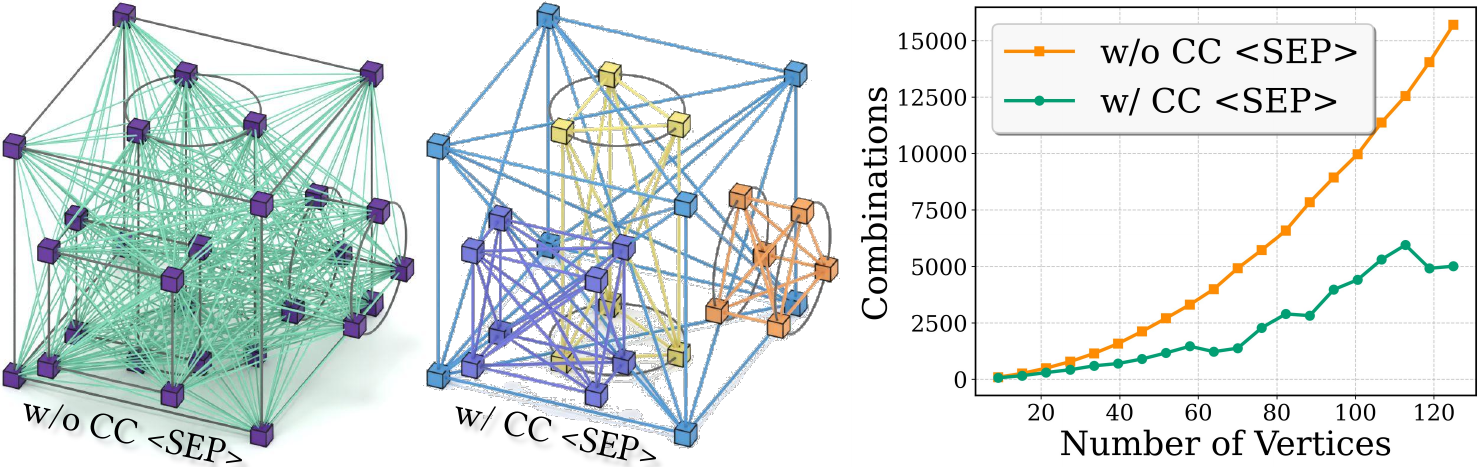}   
  \caption{Left: All pairwise connections between B-rep vertices (green). Center: Pairwise connections between vertices within connected components (different colors represent distinct subgraphs). Right: Comparison of required vertex pair computations with and without connected component tokenization, demonstrating the reduction in computational complexity.}
  \label{fig:VHP_sep}
\end{figure}

To address the quadratic complexity of processing all vertex combinations in the decoder, we decompose the B-rep vertex graph into connected components. Since edge connections only occur within these components, we can restrict the combinatorial search space from the entire B-rep to individual connected subgraphs. \re{This decomposition follows a deterministic algorithm that identifies the inherent connected subgraph structure of B-reps, where each connected component corresponds to a topologically distinct region of the model.} This decomposition is implemented by introducing \texttt{<sep>} tokens in the sequence representation to delineate component boundaries. As illustrated in Figure \ref{fig:VHP_sep}, \re{where each connected subgraph is represented with a distinct color,} this connected component separation mechanism significantly reduces the number of vertex combinations from approximately 13,000 to 5,000 for B-reps with 100 vertices, leading to improved training efficiency and effectiveness.

\paragraph{VQ-VAE for VHP} 

In the second VQ-VAE designed for VHP encoding, vertices again serve as graph nodes, but the edges are directed and include comprehensive VHP data, encompassing geometric details of curve and surface, as well as topological attributes including next-edge connectivity and inner/outer loop classification. We adopt an edge-featured graph attention mechanism \cite{monninger2023scene} to encode both geometric and topological information in the local neighborhood. The feature vectors are produced as
\begin{equation}
    \text{E}_{\text{VHP}}(\mathcal{G}^v) = \mathbf{Z}^{v} = (z^{v}_1,z^{v}_2,\ldots,z^{v}_N),
\end{equation}
where $\mathcal{G}^v$ is the directed VHP-based graph, and $z^{v}_i \in \mathbb{R}^{N_z}$ represents the learned embedding for a vertex.

Analogous to connectivity encoding, embeddings undergo RQ and are represented as $D_v$ discrete codes from the VHP-specific codebooks $\mathcal{B}^{v}$:
\begin{equation}
    \mathbf{t}^v_{i} = (t^v_{i,1}, t^v_{i,2}, \ldots, t^v_{i,D_v}) = \text{RQ}(z^v_i; \mathcal{B}^{v}, D_v).
\end{equation}

During decoding, we also employ a ResNet-based decoder to reconstruct comprehensive VHP attributes from concatenated vertex embeddings:
\begin{equation}
\hat{\mathcal{V}}_{ij} = \text{D}_{\text{VHP}} ( z_i^{v} \oplus z_j^{v} ),
\end{equation}
where $\hat{\mathcal{V}}_{ij} \in \mathbb{R}^{(N_c+1) \times N_s \times 3 + 1}$ deotes the recovered VHP information between $i$-th and $j$-th vertices.

The training of this model is guided by multiple aspects, including geometric reconstruction, next-edge prediction, inner/outer classification, and vector quantization efficiency. Specifically, the loss function is defined as:

\begin{equation}
    \mathcal{L}_{\text{VHP}} = \alpha(\mathcal{L}_{\text{geo}} + \mathcal{L}_{\text{next}}) + \beta \mathcal{L}_{\text{cls}} + \lambda \mathcal{L}_{\text{vq}},
    \label{eq:vhp_loss}
\end{equation}
where $\mathcal{L}_{\text{geo}}$ and $\mathcal{L}_{\text{next}}$ are computed using mean squared error (MSE), and $\mathcal{L}_{\text{cls}}$ employs weighted binary cross-entropy. 
The hyperparameters $\alpha$, $\beta$, and $\lambda$ control the relative contributions of different training objectives.

\subsection{Autoregressive B-rep Generation}

Our approach frames B-rep generation as a sequence generation task, with the quantized sequence structure defined in Eq. \ref{eq:sequence}. It consists of a series of vertex tokens, with predefined \texttt{<start>} and \texttt{<end>} tokens placed at the beginning and end of the sequence, respectively, and \texttt{<sep>} tokens used to separate connected components. 
With the introduction of \texttt{<sep>} tokens as described in Section \ref{sec:vq_vae_connect}, a hierarchical vertex ordering becomes necessary to correctly recover the connectivity during decoding. We implement this by sorting vertices within each connected component by ascending (z, y, x) coordinates, while ordering the connected components themselves based on the minimum (z, y, x) coordinates of their constituent vertices.
We utilize a GPT-style decoder-only Transformer to autoregressively generate these tokens, including vertex coordinates, connectivity codes, VHP codes, and custom special tokens. The connectivity and VHP codes are derived from dual VQ-VAEs, representing the indices of the corresponding embeddings in their respective codebooks. To prepare inputs for the Transformer, all tokens are converted into high-dimensional learnable embeddings.  Furthermore, discrete positional encodings are integrated to capture the relative order of tokens within the sequence, which is particularly crucial given the heterogeneous nature of our sequence components. Through multiple layers of multi-head self-attention, the Transformer is trained to iteratively predict the next token in the sequence, until meeting the \texttt{<end>} token.

Upon completion of training, the Transformer can be employed to autoregressively generate token sequences that capture the structure of B-reps. 
\re{During inference, we implement a constrained autoregressive generation strategy by applying validity masks \cite{nash2020polygen} to the Transformer's output logits.} Specifically, we enforce two key constraints: (1) vertex token predictions are restricted to their respective vocabulary ranges, and (2) delimiter tokens (\texttt{<sep>} and \texttt{<end>}) are only permitted following complete vertex descriptions (11 tokens of coordinates, connectivity and VHP). This masking mechanism helps maintain the structural integrity of the generated B-rep sequences while preserving computational efficiency.

\subsection{Detokenization}
After the Transformer predicts the tokenized sequences, we use the dual VQ-VAE decoders to construct the VHP representation based on these sequences. \re{During detokenization, the two VQ-VAEs operate sequentially. The connectivity VQ-VAE first decodes topological relationships to establish vertex connections, followed by the geometric VQ-VAE that decodes VHP features on each half-edge.}

\re{Our vertex-based approach decodes VHP features by combining each vertex feature with features from different neighboring vertices. When a vertex serves as a start point, we concatenate its features with various endpoint features to decode distinct VHPs for multiple half-edges.} The connectivity VQ-VAE establishes topological relationships between vertices, which are then materialized into curves using the geometric information encoded in VHP. \re{Each edge in the connectivity graph directly corresponds to a pair of half-edges. We decode each edge by processing features from its two endpoints, automatically grouping the resulting half-edges.}
\re{To ensure geometric alignment between half-edges sharing the same edge, we decode these half-edges from two vertex features with identical endpoints. For intermediate sample points along the edge, we compute the average position to maintain geometric consistency.} 

The cycle detection leverages the next pointers in VHP, where each vertex maintains equal in-degree and out-degree. Let $E_{in}$ and $E_{out}$ denote the sets of incoming and outgoing half-edges at a vertex. The next pointer assignment is formulated as a linear assignment problem:
\begin{equation}
\min_{\pi} \sum_{i=1}^{|E_{in}|} d(p_i, c_{\pi(i)}) \quad \text{s.t.} \quad \pi(i) \neq twin(i) \quad \forall i \in E_{in},
\end{equation}
where $p_i$ represents the next pointer sample of the $i$-th incoming edge, $c_j$ denotes the curve geometry of the $j$-th outgoing edge, $\pi$ is a permutation function that maps incoming edges to outgoing edges, and $d(\cdot,\cdot)$ measures their geometric distance. The constraint ensures that no half-edge is paired with its twin.

\re{Following cycle identification, we determine the loop-face topology by first classifying all detected loops as either outer or inner loops based on their VHP labels. We then generate initial faces by filling all outer loops using OpenCascade \cite{pyOCCT}'s n-sided filling algorithm, with the surface sample points from VHP serving as geometric constraints. For each inner loop, we compute the mean distance between its curve samples and all previously generated outer loop faces to determine its parent face.} Specifically, the parent face is chosen as the one with the minimum mean distance to the inner loop's sample points.

\subsection{Implementation Details}

For the Voronoi Half-Patch sampling strategy, we empirically set the number of curve samples $N_c$ to 6, while both the surface samples $N_s$ and next pointer samples $N_n$ are set to 4. Our framework employs Residual Vector Quantization \cite{zeghidour2021soundstream} for both the connectivity VQ-VAE and VHP VQ-VAE, with vertex feature dimensionality of 16. The quantization process utilizes 4 independent codebooks (non-shared), each containing 4096 entries. The dead code detection threshold for the exponential moving average (EMA) is set to 0.01. The encoder architectures differ in their graph convolution operations: the connectivity VQ-VAE implements SAGE \cite{hamilton2017inductive} convolution, while the VHP VQ-VAE utilizes EdgeGAT \cite{monninger2023scene}  convolution. Both encoders feature 6 convolutional layers with a hidden dimension of 1,024. The decoder follows a ResNet \cite{he2016deep} architecture comprising 4 residual blocks with a hidden dimension of 512.
For the loss function in Equation~\ref{eq:vhp_loss}, we set the hyperparameters $\alpha = 0.02$, $\beta = 1$, and $\lambda = 0.1$.
For the generative component, we adopt different Transformer configurations based on the task requirements. The unconditional generation leverages a GPT2-medium architecture with a block size of 3,072 tokens, enabling the processing of B-rep sequences containing up to 256 vertices. For conditional generation tasks, we employ a GPT2-base model.
 
We train BrepGPT's networks on eight NVIDIA RTX 4090D GPUs, using the PyTorch \cite{pytorch_2019} framework. The dual VQ-VAEs and the Transformer model are all trained with the AdamW optimizer \cite{kingma2014adam,loshchilov2019decoupled}, using a cosine annealing learning rate schedule over 300 epochs. The VQ-VAEs start with an initial learning rate of 1e-4, decaying to 1e-6, while the Transformer begins at 1e-4, gradually reducing to 1e-7. When training on the ABC dataset, the dual VQ-VAEs take one day to complete, while the Transformer decoder requires four days of training time.

\begin{figure}
\centering
  \includegraphics[width=1\linewidth]{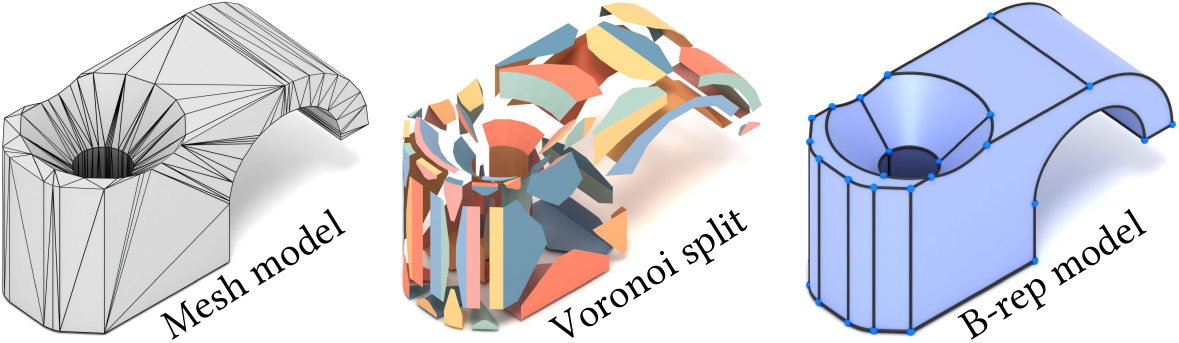}   
  \caption{\re{Visual comparison of geometric feature preservation: Our B-rep representation (right) maintains exact curve definitions, in contrast to the discretized approximation in a mesh representation (left).}}
  \label{fig:compare_mesh}
\end{figure}

\section{Experiments}
This section comprehensively evaluates BrepGPT across multiple generation paradigms: unconditional generation and controlled generation (via class labels, point clouds, images, and text). Besides these, we further demonstrate the model's capabilities in B-rep autocompletion and interpolation tasks. Ablation studies are also included to examine the impact of VQ-VAE variants and B-rep sequence ordering schemes.

\begin{figure*}[!htbp]
\centering
  \includegraphics[width=1\linewidth]{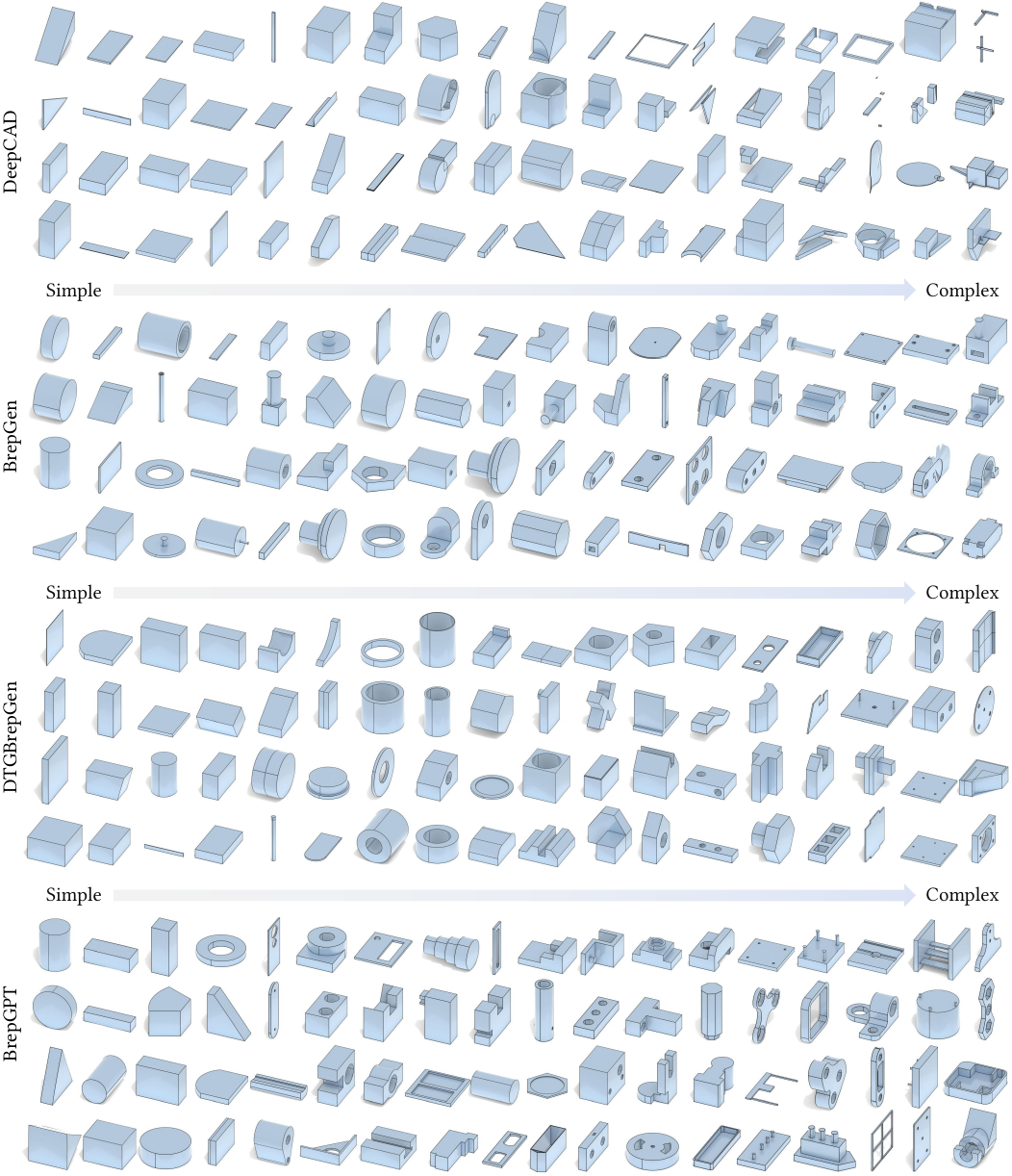}   
  \caption{Comparison of unconditional generation results against baseline methods on DeepCAD dataset, ordered by increasing vertex complexity.}
  \label{fig:compare_DeepCAD_ALL}
\end{figure*}
\begin{figure*}[!htbp]
\centering
  \includegraphics[width=1\linewidth]{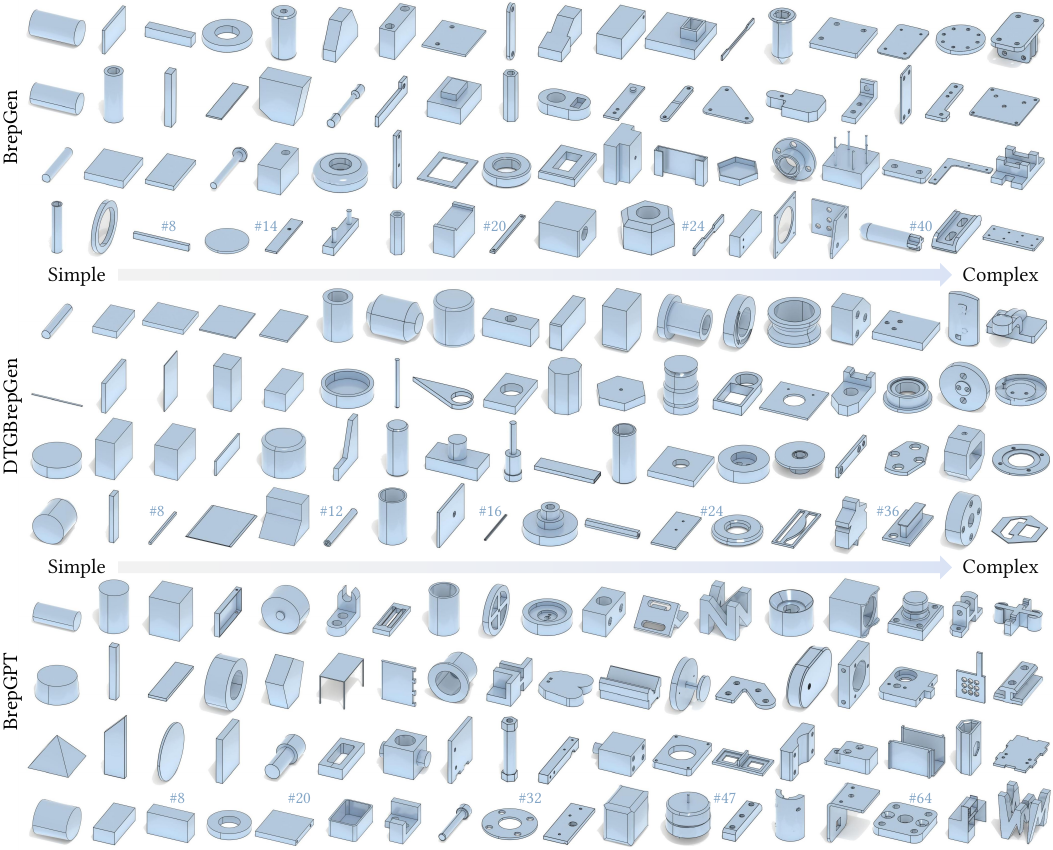}   
  \caption{Comparison of unconditional generation results against baseline methods on ABC dataset, ordered by increasing vertex complexity. \re{For the last row of each method, we annotate the vertex counts of five selected examples.}}
  \label{fig:compare_ABC_ALL}
\end{figure*}

\subsection{Experiment Setup}

\paragraph{Evaluation datasets}

Our experiments are conducted on three widely adopted datasets. The DeepCAD \cite{wu2021deepcad} and ABC \cite{koch2019abc} datasets include a wide variety of industrial parts, with DeepCAD models derived from sketch-and-extrude operations, while ABC contains more complicated and organically shaped models created using a broader set of CAD procedures. These two datasets are both employed in the unconditional generation setting. The Furniture B-rep dataset \cite{xu2024brepgen} comprises models from ten common furniture categories, exhibiting rich geometric details, and is primarily used for class-conditioned generation. The rest of the experiments, including other condition-guided tasks, B-rep autocompletion, and B-rep interpolation, are all conducted on the DeepCAD dataset. Following prior works \cite{willis2021engineering, xu2024brepgen}, we retain the original train/validation/test split from DeepCAD and exclude duplicate samples in the training set. The ABC and Furniture datasets are similarly divided in a 90\%/5\%/5\% ratio. 

\paragraph{Evaluation metrics}

In line with previous works \cite{xu2024brepgen, li2025dtgbrepgen}, we employ two categories of metrics to assess the generated results, namely \textit{distribution metrics} and \textit{CAD metrics}. \textit{Distribution metrics} measure the discrepancy between the generated set and the ground-truth set. We sample 2,000 points on the surface of each shape to calculate these metrics. \textit{CAD metrics}, meanwhile, examine the diversity and robustness of the generated CAD models.

\begin{itemize}[label=-]
    \item \textit{Distribution metrics}: \textbf{Coverage (COV)} indicates the extent to which the generated shapes cover the ground-truth set. It is calculated as the percentage of ground-truth shapes that are matched with at least one generated shape, with proximity determined by Chamfer Distance (CD). \textbf{Minimum Matching Distance (MMD)} measures how closely generated shapes correspond to ground-truth models by computing the average CD between every ground-truth model and its nearest generated match. \textbf{Jensen-Shannon Divergence (JSD)} quantifies the distribution distance between the generated set and the ground-truth set by converting the point clouds into discrete voxels and comparing their occupancy distributions.
    \item \textit{CAD metrics}: \textbf{Novel} is the proportion of generated CAD models that are unseen in the training set. \textbf{Unique} is the proportion of generated CAD models that are not repeated within the generated set. \textbf{Valid} is the proportion of generated CAD models that are watertight solid B-reps.
\end{itemize}

\begin{table}\centering
\caption{Quantitative comparison of unconditional CAD generation on DeepCAD and ABC under the \textbf{unfiltered} configuration. Both MMD and JSD are multiplied by $10^2$.}
\resizebox{\columnwidth}{!}{%
\begin{tabular}{|l|l|c|c|c|c|c|c|}
\hline
Dataset & Method & COV & MMD & JSD & Novel & Unique & Valid  \\ 
       & & $\uparrow$\smash{(\%)} & $\downarrow$ & $\downarrow$ & $\uparrow$\smash{(\%)} & $\uparrow$\smash{(\%)} & $\uparrow$\smash{(\%)} \\
\hline
\hline
\multirow{4}{*}{DeepCAD} & DeepCAD  & 65.46 & 1.294 & 1.670 & 89.8 & 89.1 & 72.6 \\
& BrepGen  & 73.87 & 1.046 & 1.286 & 99.7 & \textbf{99.2} & 61.7 \\
& DTGBrepGen  & 74.29 & 1.083 & 1.284 & \textbf{99.8} & 98.9 & \textbf{90.5} \\
& Ours     & \textbf{79.32} & \textbf{0.960} & \textbf{0.840} & {97.9} & {98.4} & 83.9 \\
\hline
\multirow{3}{*}{ABC} & BrepGen  & 69.91 & 1.185 & 0.945 & 99.6 & 99.0 & 42.9 \\ 
& DTGBrepGen  & 71.60 & 1.217& 1.142 & \textbf{99.7} & \textbf{99.1} & \textbf{64.3} \\ 
& Ours     & \textbf{73.19}  & \textbf{1.130}  & \textbf{0.933}  & {98.2} & {98.5} & 59.0 \\ 
\hline
\end{tabular}
}
\label{tab:unfiltered_performance}
\end{table}

\begin{table}[!tbp] \centering
\caption{Quantitative comparison of unconditional CAD generation on DeepCAD and ABC under the \textbf{filtered} configuration. Both MMD and JSD are multiplied by $10^2$.}
\resizebox{0.85\columnwidth}{!}{%
\begin{tabular}{|l|l|c|c|c|c|c|c|}
\hline
Dataset & Method & COV & MMD & JSD & Valid  \\ 
       & & $\uparrow$\smash{(\%)} & $\downarrow$ & $\downarrow$  & $\uparrow$\smash{(\%)} \\
\hline
\hline
\multirow{4}{*}{\shortstack{DeepCAD\\(\#vertices$\geq$12)}} & DeepCAD  & 75.44 & 1.445 & 1.286 & 60.9 \\
& BrepGen  & 69.98 & 1.292 & 1.104  & 61.5 \\
& DTGBrepGen  & 67.12 & 1.347 & 1.438 & \textbf{89.6} \\
& Ours     & \textbf{75.87} & \textbf{1.160} & \textbf{1.094} & 80.6 \\
\hline
\multirow{3}{*}{\shortstack{ABC\\(\#vertices$\geq$24)}} & BrepGen  & 67.26 & 1.319 & 1.246 & 42.4 \\ 
& DTGBrepGen  & 66.87 & 1.431 & 1.595  & \textbf{57.4} \\ 
& Ours     & \textbf{71.34}  & \textbf{1.297}  & \textbf{1.172} & \textbf{57.4} \\ 
\hline
\end{tabular}
}
\label{tab:filtered_performance}
\end{table}

\paragraph{Curve discretization validation} One key advantage of B-rep representations over mesh-based approaches is their ability to exactly represent curved features through parametric definitions, while meshes approximate curves through discrete triangular facets (Figure~\ref{fig:compare_mesh}). Although our method discretizes curves into point sequences for neural processing, this discretization is performed at high resolution to preserve the geometric precision inherent in B-rep models. To validate that our discretization strategy preserves the geometric fidelity of parametric curves, we conduct a reconstruction accuracy assessment on the ABC~\cite{koch2019abc} dataset. We randomly sample 1,000 models and evaluate the precision of our curve sampling approach by discretizing each curve with 100 uniformly distributed points, then measuring the reconstruction error against the original parametric representations. Our B-rep discretization achieves a mean error of 2.32e-5, significantly outperforming the 2.36e-3 error of equivalent mesh sampling. This validation confirms that our discretization strategy maintains the inherent geometric advantages of B-rep representations while enabling efficient sequence-based processing.

\subsection{Unconditional Generation}

\paragraph{Evaluation setting}

We evaluate BrepGPT in the unconditional setting against DTGBrepGen \cite{li2025dtgbrepgen}, BrepGen \cite{xu2024brepgen}, and DeepCAD \cite{wu2021deepcad}. DTGBrepGen and BrepGen represent state-of-the-art direct B-rep generation approaches and are applicable to both the DeepCAD \cite{wu2021deepcad} and ABC \cite{koch2019abc} datasets. In contrast, DeepCAD produces modeling sequences that are subsequently post-processed into B-reps, limiting its evaluation to the DeepCAD dataset with available construction history. 

We observe that both datasets contain a significant number of simple shapes, which fail to adequately represent the geometric and topological complexity of the B-rep representation, yet exert a strong influence on evaluation metrics. To mitigate this, we employ two experimental configurations for unconditional generation. The first (Table \ref{tab:unfiltered_performance}) adheres to the standard practice used in earlier work, retaining the full generated and reference sets. The second (Table \ref{tab:filtered_performance}) further filters out both generated and reference models with vertex counts below certain thresholds—12 vertices for DeepCAD and 24 for ABC—thereby focusing the evaluation on each method’s ability to capture the distribution of complicated CAD models. For both configurations, each method is tested over 10 evaluation runs, with metric scores averaged across runs. In each run, we randomly sample 3,000 generated shapes and 1,000 reference ground-truth shapes to compute \textit{Distribution Metrics}, while \textit{CAD Metrics} are evaluated within the set of generated models. 

\paragraph{Result comparison}

\begin{figure}[!tbp]
\centering
  \includegraphics[width=1\linewidth]{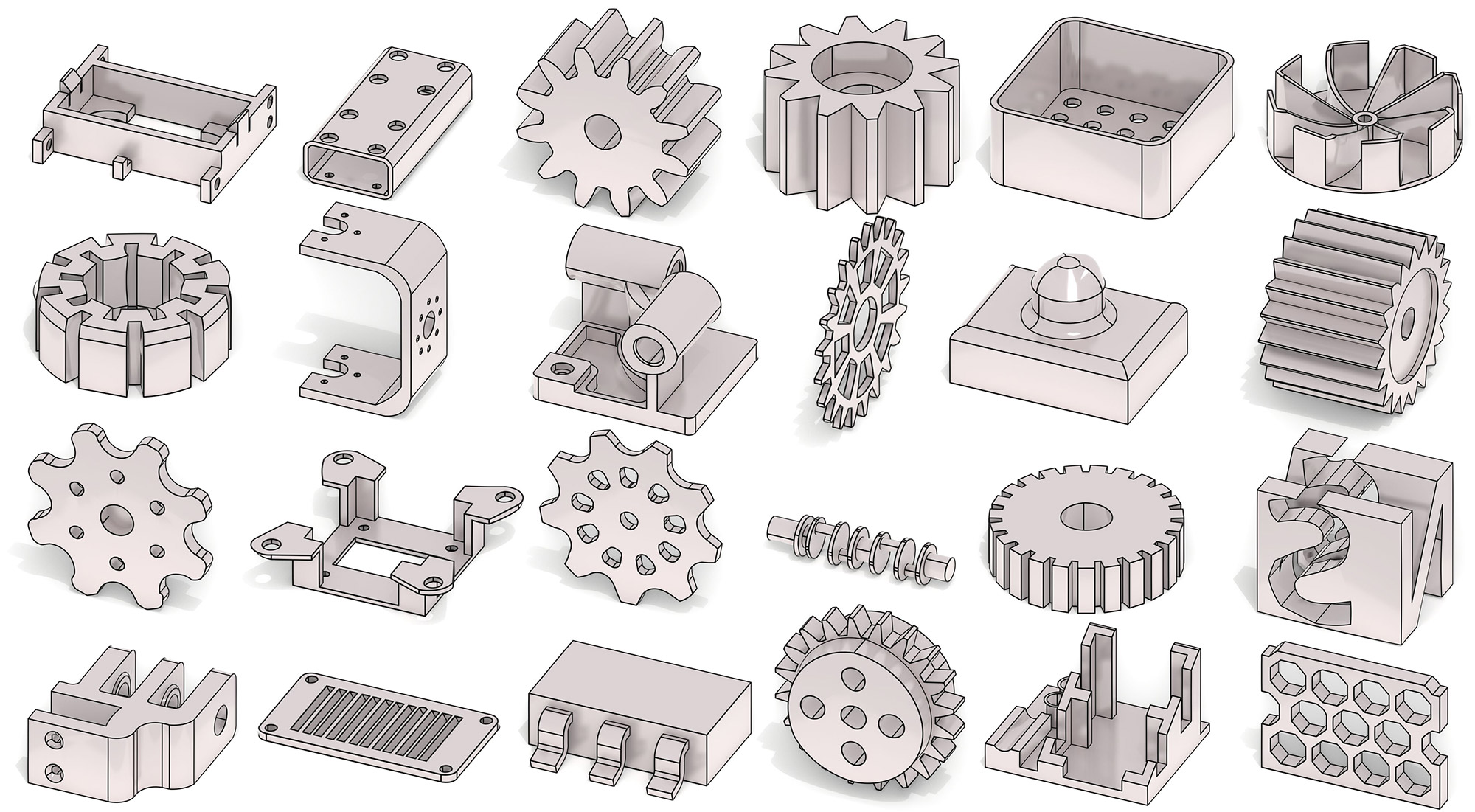}  
  \caption{Qualitative results of unconditional B-rep generation using BrepGPT, trained exclusively on complex samples (128-256 vertices) from the ABC dataset.}
  \label{fig:ABC_128_256}
\end{figure}

\begin{figure}[!htbp]
\centering
  \includegraphics[width=1\linewidth]{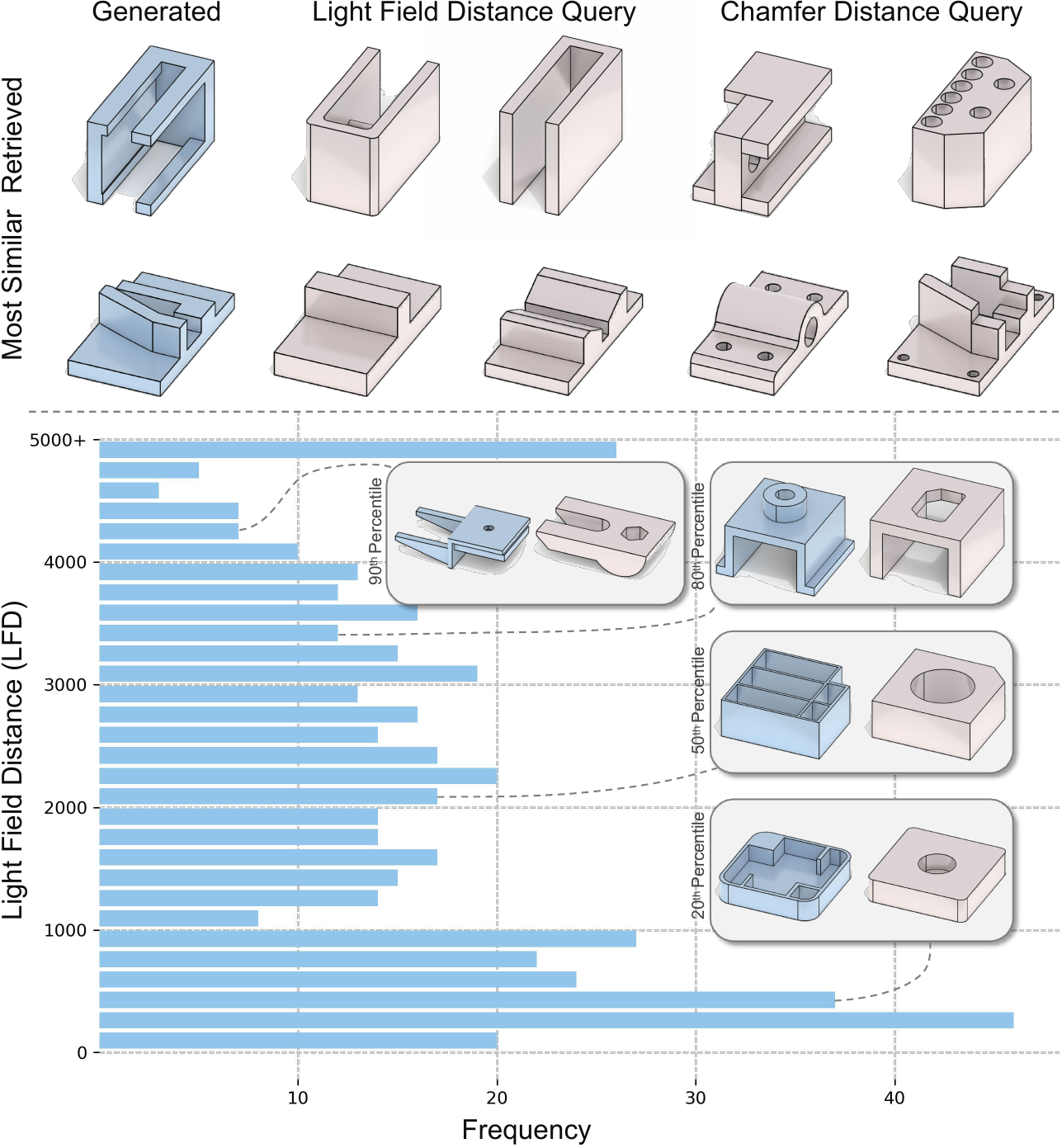}   
  \caption{Novelty analysis of generated B-rep models. Top: Nearest neighbor retrieval from the training set (pink) for generated shapes (blue) using Light Field Distance (LFD) and Chamfer Distance metrics. Bottom: Distribution of LFD between 500 randomly sampled generations (blue) and their closest training counterparts (pink). }
  \label{fig:LFD}
\end{figure}

As shown in Table \ref{tab:unfiltered_performance} and Table \ref{tab:filtered_performance}, BrepGPT consistently outperforms other state-of-the-art methods in terms of COV, MMD, and JSD across both configurations and datasets. These results demonstrate the superior ability of our method to generate diverse and structurally complex B-rep models. We can see that the filtered configuration introduces noticeable effects. DeepCAD achieves a notable improvement in COV but suffers a substantial drop in Valid. While DTGBrepGen maintains a relatively high Valid score, its performance on other metrics shows limitations when handling complex models. In contrast, BrepGPT slightly lags behind DTGBrepGen in Valid, but achieves a better trade-off between the quality and validity of generated results, particularly under the filtered setting. 
As for Novel and Unique metrics, BrepGPT's relatively lower scores compared to diffusion-based methods could be attributed to the discrete vocabulary space induced by vector quantization.
However, these two metrics should be interpreted with caution as they treat both minor variations and quality issues (such as noisy outputs and structural defects) as \textit{novel} or \textit{unique} instances. These two metrics are omitted in Table \ref{tab:filtered_performance} due to minimal variation.
To provide a more rigorous assessment, we evaluate BrepGPT's generative capabilities using Light Field Distance (LFD) distributions and nearest neighbor analysis, following prior works \cite{hui2022wavelet,siddiqui2024meshgpt,xu2024brepgen}.
As illustrated in Figure \ref{fig:LFD}, the top section shows representative generated models alongside their most similar ground-truth shapes retrieved using Light Field Distance (LFD) and Chamfer Distance (CD). The lower part in Figure \ref{fig:LFD} visualizes the LFD distribution between 500 generated models and their closest matches in the training set. These results suggest that BrepGPT is capable of generating novel yet plausible shapes that differ from the training data.

In Figure \ref{fig:compare_DeepCAD_ALL} and Figure \ref{fig:compare_ABC_ALL}, we qualitatively illustrate how each method performs on the DeepCAD and ABC datasets, respectively. \re{The displayed models are randomly sampled from each method and arranged in ascending order of vertex complexity.} As complexity grows, DeepCAD tends to produce diverse but often implausible shapes, aligning with its high COV yet lower scores on other metrics in Table \ref{tab:filtered_performance}. While other competitors generally maintain coherent structures under high complexity, \re{BrepGPT generates geometries with higher structural complexity. As shown in Figure \ref{fig:compare_ABC_ALL}, vertex count annotations indicate that BrepGPT generates samples with over 60 vertices in high-complexity regions (rightmost examples), compared to approximately 40 vertices for other methods.} To more thoroughly examine the effectiveness of our method on complex shapes, we retrain BrepGPT solely on complex ABC samples \re{for self-validation of our method's capabilities on these challenging cases.} A gallery of generated results is shown in Figure \ref{fig:ABC_128_256}, where BrepGPT synthesizes a diverse array of complex industrial components characterized by valid topological connections and high-fidelity local geometric details.

\begin{figure}
\centering
  \includegraphics[width=1\linewidth]{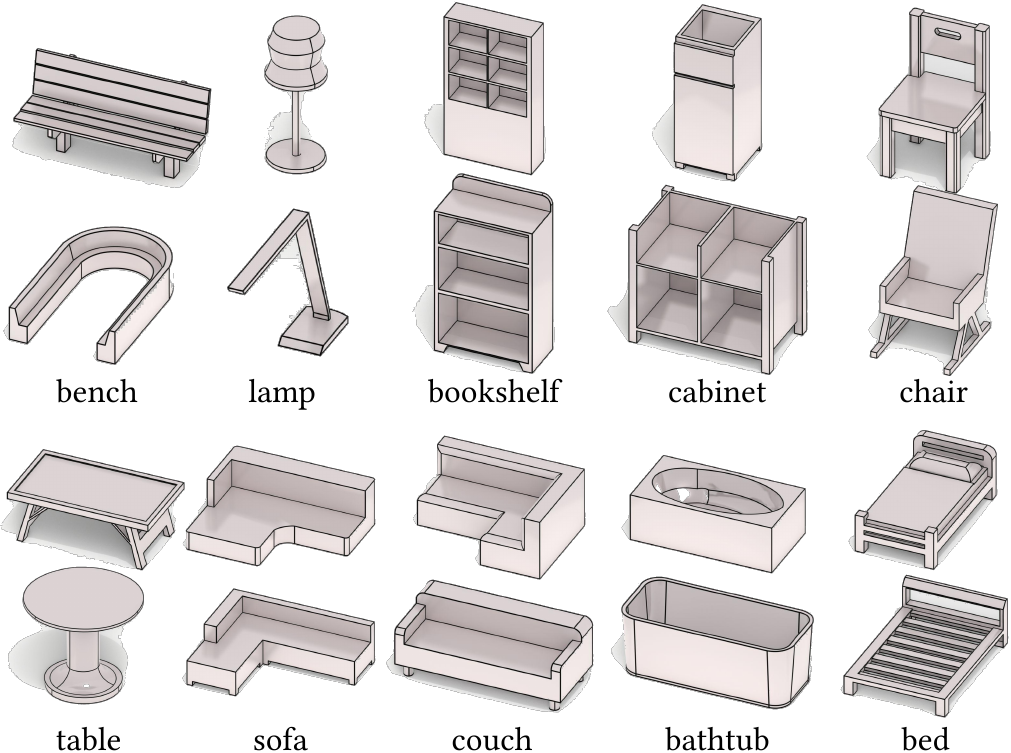}   
  \caption{Class-conditioned B-rep generation results on the Furniture dataset.}
  \label{fig:furniture}
\end{figure}

\begin{figure*}[!htbp]
\centering
  \includegraphics[width=1\linewidth]{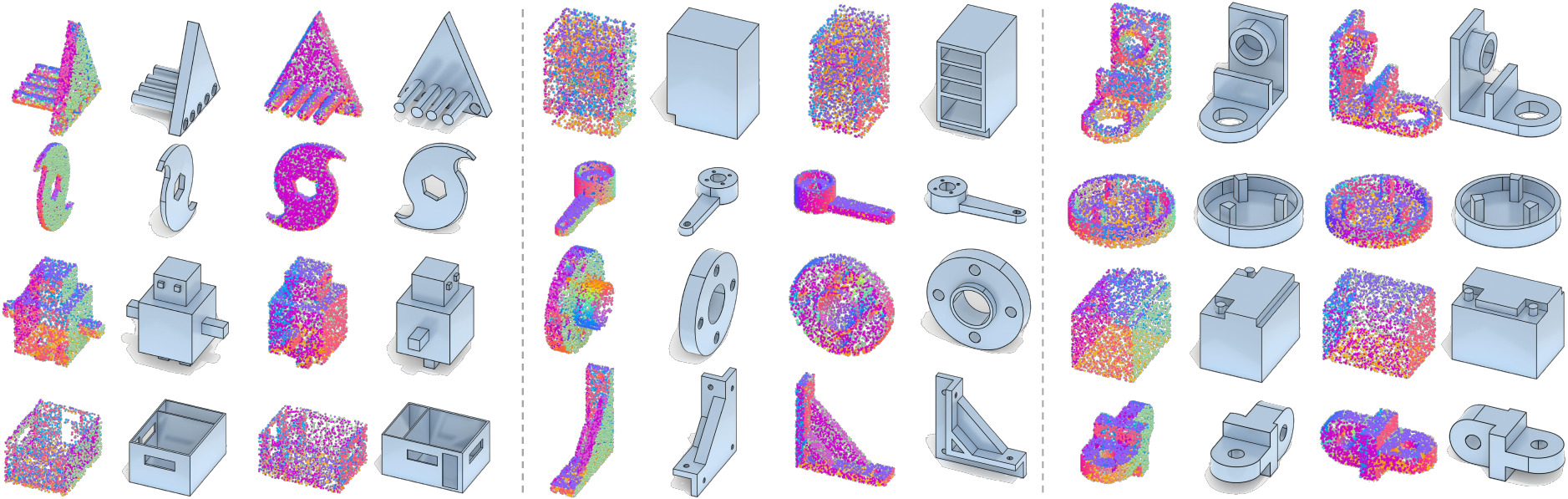}   
  \caption{Point cloud-conditioned B-rep generation results. For each example, we display the input point cloud and its generated B-rep model from two views.}
  \label{fig:pcl_cond}
\end{figure*}
\begin{figure}[!htbp]
\centering
  \includegraphics[width=1\linewidth]{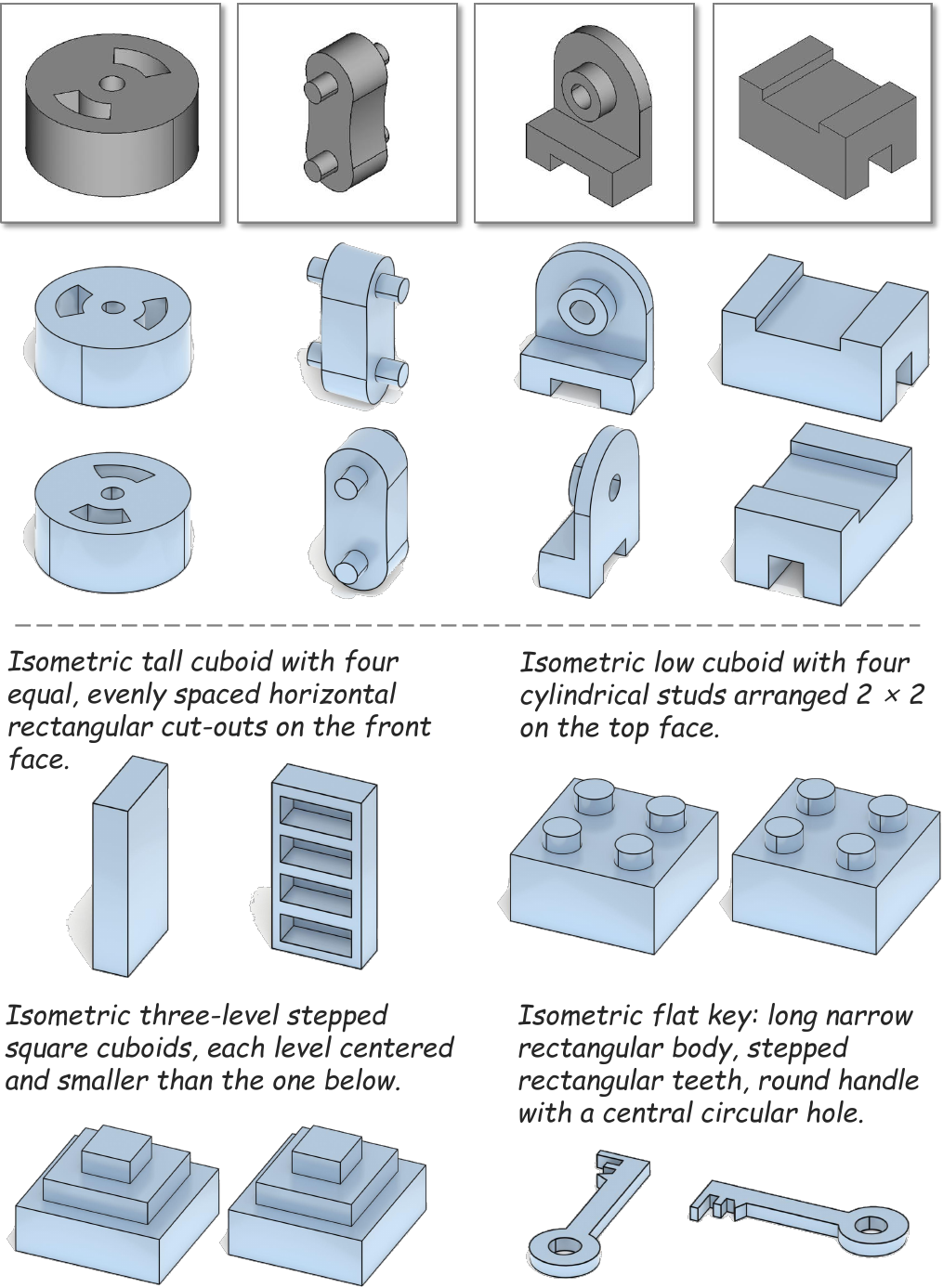}   
  \caption{Results of B-rep generation conditioned on a single image (top) versus text input (bottom). The generated B-rep models are displayed from two different viewpoints.}
  \label{fig:text_img}
\end{figure}

\begin{figure}[!htbp]
\centering
  \includegraphics[width=1\linewidth]{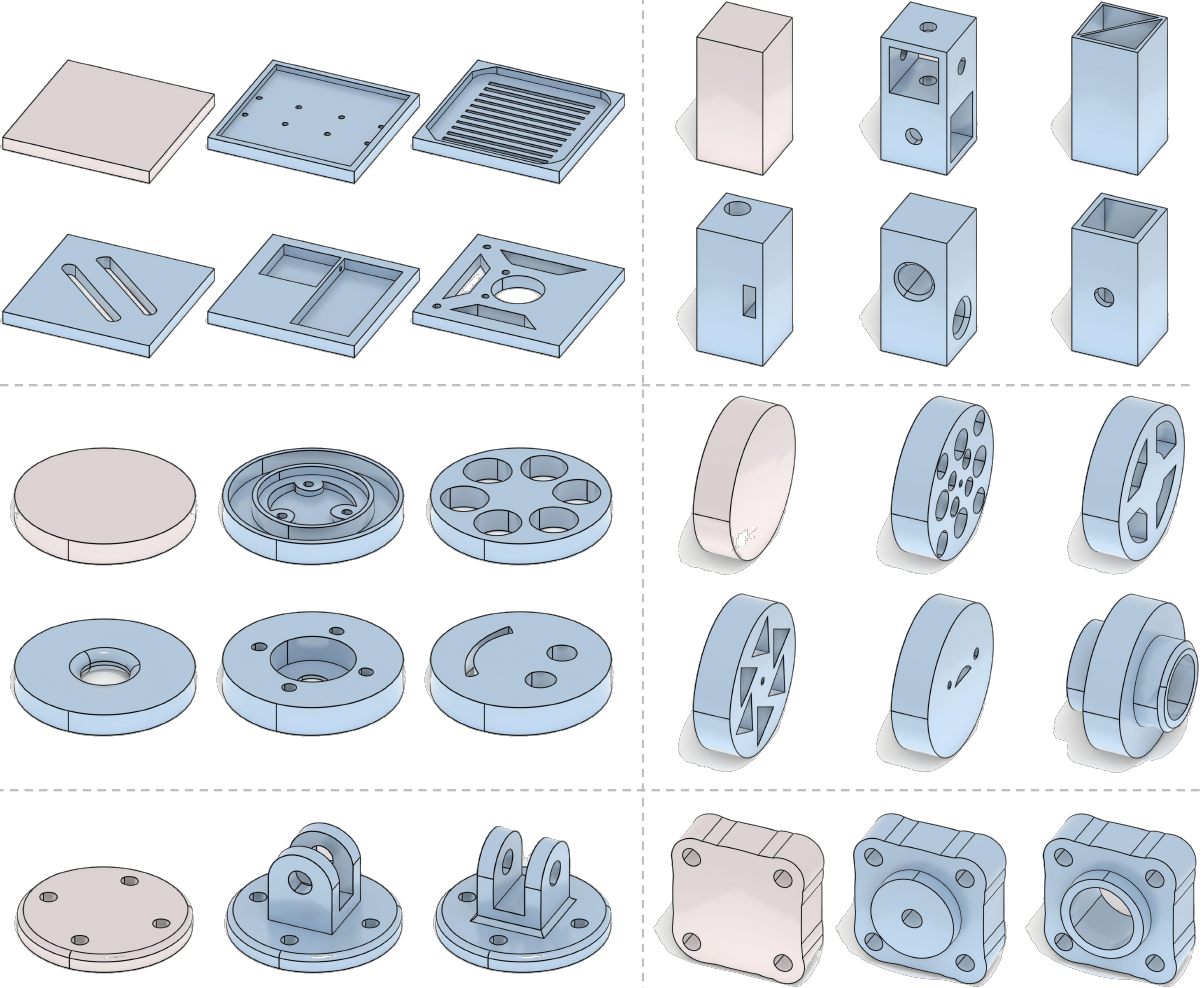}   
  \caption{Results of automatic B-rep completion. For each example, the pink B-rep represents the partial input, while the blue models demonstrate various completion alternatives generated by our method.}
  \label{fig:autocomplete}
\end{figure}

\begin{figure}[!htbp]
\centering
  \includegraphics[width=1\linewidth]{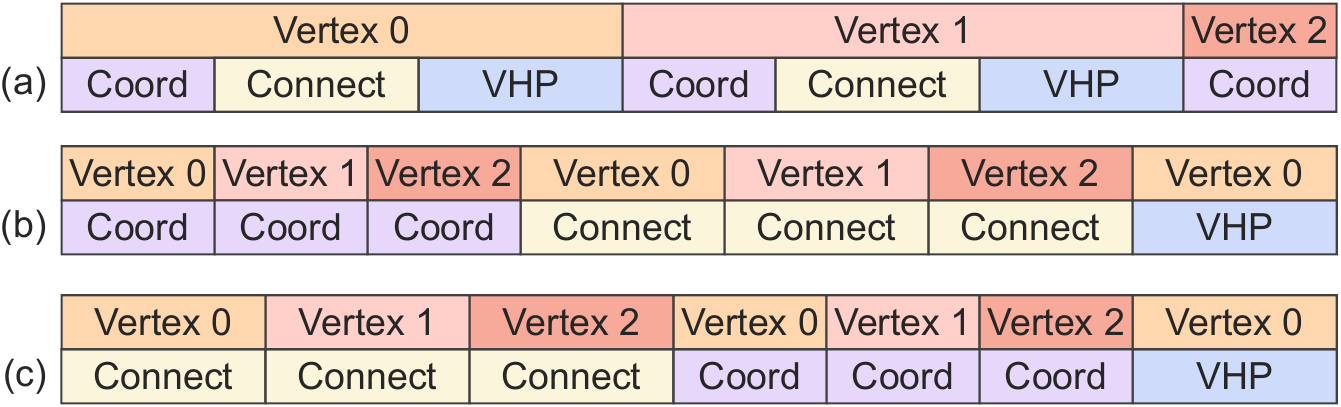}   
  \caption{Variants of token ordering. (a) Vertex-based: alternating coordinate, connectivity, and VHP tokens per vertex. (b) Coordinate-first: sequential arrangement of all coordinate tokens followed by connectivity and VHP tokens. (c) Topology-first: prioritizing connectivity tokens, followed by coordinate and VHP tokens.}
  \label{fig:seq_order}
\end{figure}

\begin{figure*}[!htbp]
\centering
  \includegraphics[width=1\linewidth]{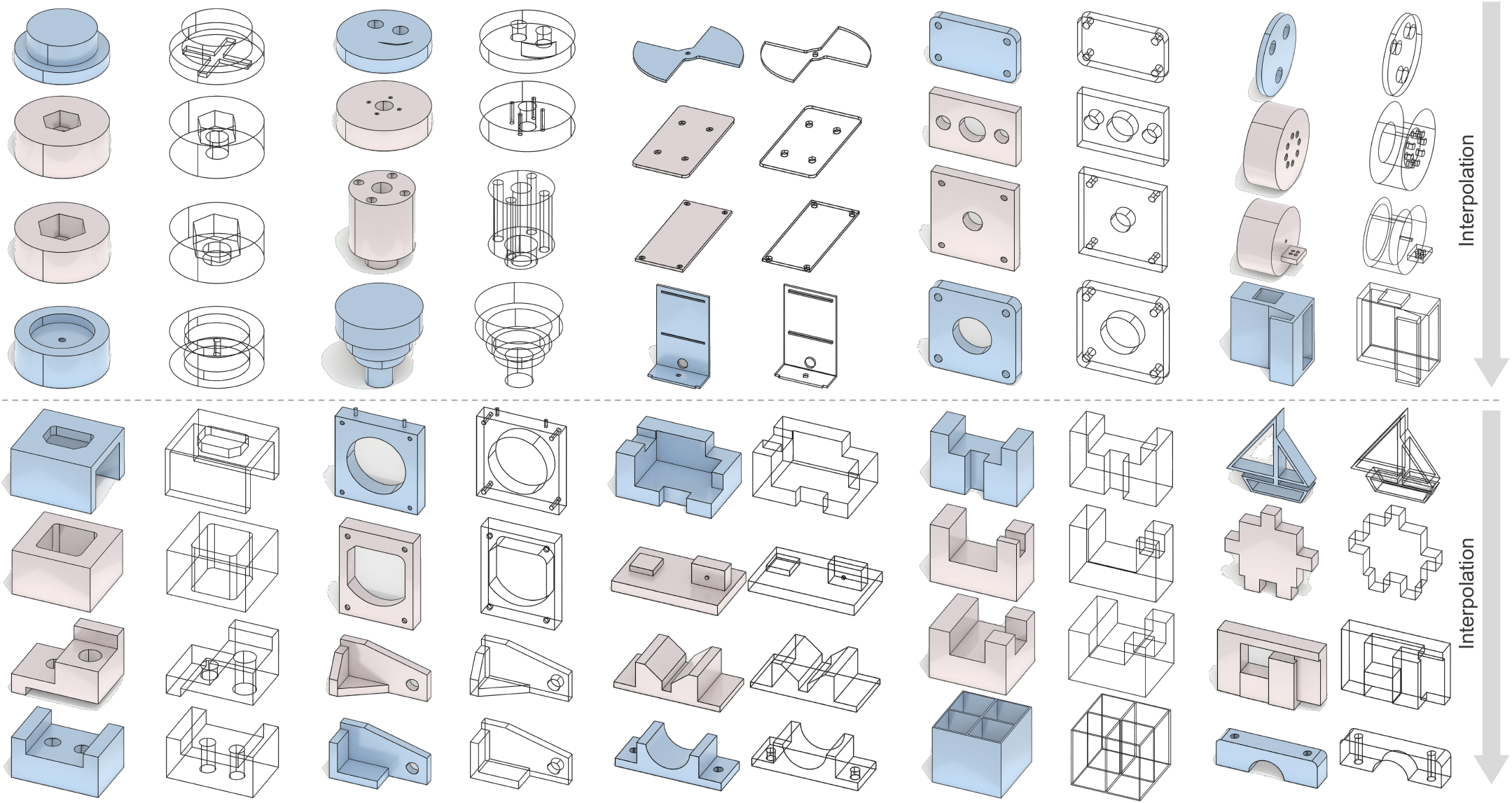}   
  \caption{Demonstration of B-rep shape interpolation. Two distinct B-reps (blue) are generated from point cloud conditions, with intermediate states (pink) obtained through linear interpolation in the latent space. Corresponding wireframe views reveal the detailed evolution of internal topological structures.}
  \label{fig:interpolation}
\end{figure*}

\subsection{Controllable Generation}
Given the autoregressive nature of our GPT-based architecture, we can leverage various established conditional generation techniques from language modeling and adapt them to B-rep generation. 
This enables a wide spectrum of controlled generation scenarios, ranging from class-conditional synthesis to multi-modal generation guided by point clouds, images, or text descriptions. Furthermore, the sequential modeling capability allows for sophisticated B-rep manipulation tasks such as autocompletion and interpolation.

\paragraph{Class-conditioned generation}
We evaluate our approach on a benchmark furniture dataset \cite{xu2024brepgen} comprising 10 categories. By substituting standard start tokens with learned category embeddings, we enable controlled generation of category-specific B-rep models. Our implementation accommodates B-rep structures with up to 256 vertices, facilitating the synthesis of geometrically complex assemblies, such as benches and beds composed of multiple planar elements and structural members. The qualitative results presented in Figure \ref{fig:furniture} demonstrate the framework's capability to generate category-specific architectural features while maintaining geometric and topological validity across diverse furniture types.

\paragraph{Point cloud-conditioned generation}
For conditional generation from point clouds, we adopt an approach inspired by \cite{chen2024meshanything}. For each input shape, we sample 4,096 points along with their surface normals. These points are encoded using a pre-trained point encoder \cite{zhao2023michelangelo}, which was trained on the ShapeNet dataset \cite{chang2015shapenet} and remains frozen during our training process. The encoder maps the input point cloud into a sequence of 257 feature tokens (each with a 768-dimensional embedding). By prepending these conditioning tokens to the generation sequence, our model learns to synthesize B-rep models that align with the input point cloud geometry. \re{We evaluate this capability on the DeepCAD \cite{wu2021deepcad} dataset.} Figure \ref{fig:pcl_cond} demonstrates the effectiveness of our approach, showing generated B-reps that match the input point clouds from two viewpoints.

\paragraph{Image and text-conditioned generation}
Building upon our point cloud-conditioned model, we extend our framework to support both image and text inputs for B-rep geometry generation. For image-to-B-rep generation, we utilize Rodin Gen-1.5 \cite{DeemosRodin2024} to obtain an intermediate mesh representation, which is then sampled to create point clouds. These point clouds serve as conditioning inputs for our point cloud-conditioned B-rep generative model. For text-to-B-rep generation, we employ ChatGPT-4o \cite{gpt4o_2025} to synthesize intermediate images from textual descriptions, which then follow the same image-to-B-rep pipeline. \re{These experiments are conducted on the DeepCAD \cite{wu2021deepcad} dataset.} As demonstrated in Figure \ref{fig:text_img}, BrepGPT successfully generates B-rep models that faithfully align with both input images and textual descriptions.

\begin{table}[!tbp]
\centering
\caption{Ablation study on VQ-VAE variants. The SAGE rows show F1 scores for connections, while EdgeGAT rows present both connection F1 scores and MSE of VHP in the format "F1/MSE". All values are multiplied by $10^2$.}
\label{tab:vqvae_ablation}
\resizebox{0.95\columnwidth}{!}{%
\begin{tabular}{|c|c|c|c|c|c|}
\hline
\multirow{2}{*}{Encoder} & \multirow{2}{*}{\begin{tabular}[c]{@{}c@{}}Codebook\\Size\end{tabular}} & \multicolumn{4}{c|}{Number of Quantizers} \\
\cline{3-6}

& & 1 & 2 & 4 & 8 \\
\hline
\hline
\multirow{4}{*}{SAGE} & 1,024 & 94.64 & 98.13 & 99.86 & 99.90 \\
& 2,048 & 95.35 & 98.22 & 99.95 & 99.96 \\
& 4,096 & 95.88 & 98.48 & 99.99 & 99.99 \\
& 8,192 & 95.89 & 98.73 & 99.99 & 99.99 \\
\hline
\multirow{4}{*}{EdgeGAT} & 1,024 & 94.4/1.37 & 96.4/1.14 & 98.1/0.78 & 98.3/0.85 \\
& 2,048 & 95.1/1.31 & 96.3/1.16 & 98.3/0.81 & 98.6/0.83 \\
& 4,096 & 95.7/1.29 & 96.6/1.11 & 98.1/0.88 & 98.5/0.81 \\
& 8,192 & 95.9/1.29 & 96.7/1.04 & 98.6/0.83 & 98.6/0.75 \\
\hline
\end{tabular}
}
\end{table}

\paragraph{B-rep autocompletion}
Our hierarchical vertex ordering scheme, which sorts vertices both within and across connected components by (z, y, x) coordinates, naturally facilitates progressive B-rep completion. We evaluate this capability on the DeepCAD dataset. While arbitrary truncation vertices are theoretically possible, we obtain partial B-rep inputs by segmenting at component interfaces from complete models to preserve partial topological validity. As shown in Figure \ref{fig:autocomplete}, our method successfully generates diverse and structurally valid completions. The top two rows demonstrate that simple partial inputs, such as basic geometric primitives, lead to multiple plausible solutions with significant geometric variations. Additionally, the bottom row shows that more complex partial B-reps can constrain the solution space, resulting in completions that maintain consistency with the input's structural features while still offering meaningful variations in the generated portions. \re{Nevertheless, this approach has certain limits in practice, as the input format constraints and subgraph-based design restrict the flexibility of arbitrary user-specified partial geometries.}

\paragraph{B-rep interpolation}
Shape interpolation for B-rep and other procedural 3D models remains challenging due to their inherent sequential nature. Unlike point clouds or signed distance fields (SDFs), procedural models comprise variable-length sequences of modeling operations, making direct correspondence mapping between shapes with different component counts infeasible. Even when correspondences can be established, maintaining topological validity during interpolation poses significant challenges. We address this limitation by introducing a novel interpolation approach that leverages fixed-length latent features as conditioning signals. By linearly interpolating between the point cloud features of two input shapes in our conditional framework, we can guide the autoregressive generation of B-rep sequences. This approach combines the advantages of continuous, fixed-dimensional latent spaces with the flexibility of variable-length autoregressive generation. As demonstrated in Figure~\ref{fig:interpolation}, BrepGPT achieves smooth interpolation between geometrically similar shapes, as shown in the leftmost example of the top row. For cases with significant topological and geometric differences, such as the rightmost in the top row, while perfect smoothness is not guaranteed, the autoregressive nature of our model ensures the generation of valid B-rep models throughout the interpolation sequence.

\subsection{Ablation Study}
\paragraph{VQ-VAE variants}
We investigated the impact of different codebook parameters for both VQ-VAE frameworks and examined architectural variants solely for the connectivity encoder, as SAGE's aggregating mechanism operates on graph topology rather than edge-based geometric features. Table \ref{tab:vqvae_ablation} presents the quantitative results. The experiments demonstrate that increasing either the codebook size or the number of quantizers leads to improved prediction accuracy, with the latter showing more substantial gains. Notably, the SAGE convolution consistently outperforms EdgeGAT convolution in the encoder architecture for connectivity encoding. Based on these findings and considering the trade-off between accuracy and computational efficiency, we adopted SAGE convolution for the connectivity VQ-VAE encoder while retaining EdgeGAT convolution for the VHP component, with a codebook size of 4,096 and 4 quantizers as our default configuration.

\begin{table}[!tbp]\centering
\caption{Ablation study on token sequence ordering.}
\resizebox{0.8\columnwidth}{!}{%
\begin{tabular}{|l|c|c|c|c|}
\hline
Method & COV$\uparrow$(\%) & MMD$\downarrow$ & JSD$\downarrow$ & Valid$\uparrow$(\%)  \\ 
\hline
\hline
Vertex-based     & {79.32} & {0.960} & \textbf{0.840} & 83.90 \\
Coordinate-first  & \textbf{79.60} & 0.989 & 0.848 & 83.38 \\
Topology-first  & 79.10 & \textbf{0.959} & 0.855 & \textbf{84.24} \\
\hline
\end{tabular}
}
\label{tab:ablation_seq_order}
\end{table}

\paragraph{Sequence ordering} \label{sec:seq_order}
We investigate different token arrangement strategies for our vertex-based sequence representation, which comprises coordinate, connectivity, and VHP tokens. 
Beyond the interleaved arrangement where x-y-z coordinates alternate with connectivity and VHP information for each vertex, we explore coordinate-first and connectivity-first sequencing patterns shown in Figure \ref{fig:seq_order}.
Our ablation study on the DeepCAD dataset reveals that these different token arrangements yield comparable generation quality, as reported in Table \ref{tab:ablation_seq_order}. While the quantitative results suggest minimal impact on overall performance, we observe that each arrangement exhibits distinct characteristics advantageous for specific applications. The interleaved arrangement facilitates B-rep autocompletion task.
The coordinate-first arrangement enables diverse topological variations while maintaining identical vertex positions, whereas the connectivity-first arrangement allows geometric variations while preserving topological structure. 
Given its natural representation of geometric entities, we adopt the interleaved arrangement as our default configuration.

\section{Limitations and Future Work}
Despite introducing a novel decoder-only architecture for B-rep generation, BrepGPT exhibits several limitations that merit further investigation.
First, while invalid masking is utilized to encourage valid outputs, the lack of explicit constraints during generation can lead to invalid B-reps, such as those with isolated vertices or self-intersecting faces. Incorporating a validation step with rollback functionality \cite{pun2025legogpt} could mitigate this issue.
Second, while our vertex-based encoding scheme effectively compresses B-rep structures into compact token sequences, the model still exhibits the characteristic limitations of autoregressive architectures when handling longer sequences. Future work could explore complementary encoding strategies, such as enhanced codebook designs or advanced VQ-VAE architectures, to further optimize sequence representation. Third, the discretization inherent in Vector Quantization introduces precision loss, particularly evident in geometrically complex B-reps. Since VHP representation remains independent of specific network architectures, future research could explore diffusion-based generation methods or hybrid approaches combining autoregressive and diffusion techniques \cite{li2024autoregressive} to enhance the geometric precision of generated B-reps.

\section{Conclusion}
We present BrepGPT, the first decoder-only architecture for autoregressive B-rep generation. The core of our approach is the Voronoi Half-Patch (VHP) representation, which decomposes B-reps into unified sequences based on half-edge data structures. The VHP representation unifies geometric attributes and topological structures within each local decomposition unit, enabling a more coherent representation of B-rep manifolds. Using dual VQ-VAEs to encode vertex connectivity and VHP information, we transform B-reps into compact vertex-based token sequences. A GPT-style decoder learns these sequences for autoregressive synthesis. BrepGPT demonstrates strong performance in unconditional generation, conditional tasks (class, image, and text-guided synthesis), as well as B-rep autocompletion and interpolation. 
We believe the BrepGPT architecture demonstrates how autoregressive generation can be effectively applied to B-rep modeling, while the VHP representation introduces a learning-friendly B-rep formulation that could benefit future B-rep generation and analysis.

\begin{acks}
The authors thank the reviewers for their constructive comments. This work was supported in part by the Key Project of the National Natural Science Foundation of China (12494550, 12494553), the NSFC-FDCT Joint Project (62461160259), the Beijing Natural Science Foundation (Z240002), the Strategic Priority Research Program of the Chinese Academy of Sciences (XDB0640000 and XDB0640200), the Guangdong Basic and Applied Basic Research Foundation (2023B-1515120026), and King Abdullah University of Science and Technology (KAUST) - Center of Excellence for Generative AI (5940).
\end{acks}

\bibliographystyle{ACM-Reference-Format}
\bibliography{ref}

\end{document}